\documentclass[final,5p,times,twocolumn]{elsarticle}

\usepackage{amssymb}
\usepackage[fleqn]{amsmath}
\usepackage{algorithm}
\usepackage{algpseudocode}

\usepackage{booktabs, threeparttable}
\usepackage{caption}

\usepackage[labelfont=bf,justification=raggedright]{caption}
\usepackage{amsthm}

\usepackage{url}

\usepackage{color}

\journal{Neurocomputing}

\begin{document}

\begin{frontmatter}


 \title{Empirical curvelet based Fully Convolutional Network for supervised texture image segmentation}

 \author[1]{Yuan Huang\corref{cor1}}
  \ead{huangyuan@buaa.edu.cn}
  \address[1]{Image Processing Center, Beihang University, Beijing 100083,
China}
  
 \author[1]{Fugen~Zhou}
  \ead{zhfugen@buaa.edu.cn}
 \author[2]{~J\'er\^ome~Gilles}
 
 \ead{jgilles@sdsu.edu}
 \cortext[cor1]{Corresponding author}
 \address[2]{Department of Mathematics and Statistics, San Diego State University, 5500 Campanile Dr, San Diego, CA, 92182, USA}

\setlength{\mathindent}{0pt}

\noindent





\begin{abstract}
In this paper, we propose a new approach to perform supervised texture classification/segmentation. The proposed idea is to feed a Fully Convolutional Network with specific texture descriptors. These texture features are extracted from images by using an empirical curvelet transform. We propose a method to build a unique empirical curvelet filter bank adapted to a given dictionary of textures. We then show that the output of these filters can be used to build efficient texture descriptors utilized to finally feed deep learning networks. Our approach is finally evaluated on several datasets and compare the results to various state-of-the-art algorithms and show that the proposed method dramatically outperform all existing ones.
\end{abstract}


\end{frontmatter}

\noindent
\setlength{\mathindent}{0cm}

\section{Introduction}
{T}{exture} analysis is an important aspect of computer vision, it is involved in a wide variety of applications ranging from medical imaging (tumor detection) \cite{tumor}, 
camouflaged target detection \cite{ctd1,ctd2} or 
microscopy imaging (analysis of molecule alignments) \cite{micro,micro2}, \ldots However, texture segmentation and/or classification is still an open and challenging problem in 
image processing and computer vision due to the fact that textures can have a huge variability making them difficult to characterize. Finding 
a mathematical model to textures is very hard because of such variability. The absence of a general texture model explains why texture analysis is still a particularly challenging 
problem nowadays. 

Algorithms developed for texture classification/segmentation purposes are 
based on two main steps: first the extraction of texture features whose aim is to characterize textures; second a classifying step which uses the information contained in the 
features to distinguish between the different textures. Two main philosophies can be used to perform the second step: unsupervised and supervised. The unsupervised case does not use any 
prior 
knowledge based on any learning technique, while the supervised case uses a known dictionary of textures to learn how to distinguish them from each other. The 
unsupervised 
case is usually very challenging and 
in this paper we focus on the supervised case. It is worth to mention that the segmentation can be performed at the pixel level or at a ``superpixel'' level (i.e. it is a 
sursegmentation)
\cite{randomwalk,superpix}. In this paper, we intend to perform segmentation at the pixel level by introducing the use of data-driven wavelets to create efficient texture 
descriptors which will then be used to feed a neural network to perform the classification. Sections~\ref{sec:textdes} hereafter and \ref{sec:class} review state-of-the-art 
methods regarding texture descriptors and data classifiers, respectively. We also refer 
the reader to the review article \cite{Liu2019} for a recent panorama of advances in the field of texture analysis.

\subsection{Texture descriptors}\label{sec:textdes}
Several approaches were proposed in the literature to design texture features:
Tamura local descriptors \cite{Tamura,Li2019}, Haralick descriptors \cite{Haralick} based on the Gray Level Co-occurrence Matrix (GLCM) \cite{co-occ,Subudhi2018,Zhao2018}, Fractal Dimension \cite{Soares2018}, Structure Tensor \cite{Kunina2017}, Local Binary Pattern (LBP) \cite{LBP,Wang2018,Banerjee2018,Dong2018,Kazak2018,Chen2018}, Laws' Texture Energy \cite{Wardhani2018}, Hermite transform \cite{Hermite}, Complex Networks \cite{Ribas} or 
Markov Random Fields (MRF) \cite{UTSMRF}. It is generally accepted within the community that scales and orientations are important properties to characterize textures. Therefore, 
techniques based on linear filtering \cite{Lfilter,Ye2018,Song2018,Song2018a}, such as Gabor filters \cite{Gabor1,Gabor2,Kim2018,Gao2019} are widely used in the literature because of their excellent selectivity in both 
scales and orientations. The wavelet transform and its variants \cite{wavelet1,wavelet2,wavelet3,Subudhi2019} were also used since they provide more flexibility regarding the chosen basis 
functions while keeping excellent selectivity in scales and orientations. 

In the last decades, adaptive, i.e. data driven, decomposition techniques like Empirical Mode 
Decomposition 
(EMD) \cite{EMD} had received a lot of attention and reached great successes in signal and image processing to extract low-level features \cite{Yang2018}. However, the EMD method is a purely 
algorithmic method and lacks mathematical foundations, moreover the interpretation of its generalization in 2D is more challenging. More recently, in \cite{EWT1D}, the author proposed to 
build an adaptive wavelet transform, called Empirical Wavelet Transform (EWT) to leverage such problems. This transform has been extended to 2D for image processing purposes in 
\cite{EWT2D}. In 
\cite{UTSEWT}, the authors investigate the use of the EWT to perform unsupervised texture segmentation and showed that data-driven wavelet achieve 
significant better segmentation results than any other wavelet based descriptors. 

\subsection{Data classification}\label{sec:class}
The obtained texture descriptors are then employed to feed a classifier which will assign one texture class to each pixel. As 
mentioned above two approaches can be investigated: unsupervised or supervised classification/segmentation. In the unsupervised case, the algorithm only uses the current 
available information to decide to which texture a pixel belongs to. For instance, in \cite{AKBULUT2018494}, the authors combine color and texture descriptors to feed a Mean-shift 
clustering algorithm \cite{Meanshift}; in \cite{Pustelnik,Pascal2018} the authors use wavelets to characterize the regularity of textures to drive a segmentation algorithm based on total 
variation. The output of 
DOOG (Difference Of Offset Gaussians) filters were used to feed a neutrosophic clustering algorithm in \cite{neutrosophic}. In \cite{LLIF}, the authors propose a new approach 
consisting in finding the best features to represent a certain set of textures in the absence of annotations and then use a Mumford-Shah model to perform the final 
segmentation.
Among the supervised approaches, the learning step can be performed using two main strategies: directly using an annotated dataset or using some complimentary prior information. 
For instance, 
in \cite{enermin,HOE1,HOE2}, the authors propose similar algorithms based on minimizing high-order energies guided by saliency maps and external user interactions. Such algorithms 
perform poorly with textures since saliency maps are not efficient texture descriptors. The authors of \cite{Ustyuzhaninov} train a neural network to recognize textural patches to perform the segmentation task. In \cite{randomwalkmarkov}, interactively driven Markov random walks are used to 
perform the segmentation. Deep learning/neural network techniques had become very popular this last decade to tackle many segmentation/classification problems like salient 
object detection \cite{Wang_2018_CVPR,videosaliency}, semantic segmentation \cite{pyramid,Context}, hyperparameter optimization \cite{Hyperparameter},\ldots

For the specific task of texture segmentation/classification, several deep learning architectures were proposed in the literature. The authors of \cite{Salhi2018} combine several texture features to train a Kohonen Network.
In \cite{MBLearning}, the authors develop an algorithm using learnable convolutional features to perform texture segmentation while a Convolutional Neural Network (CNN) was used in 
\cite{CNN,CNN2,Andrearczyk2018,Xue2018,Fu2018} to build descriptors to classify textures. In \cite{LSTM}, a Long Short-Term Memory (LSTM) based Recurrent Neural Network (RNN) was also 
proposed to extract texture features and perform supervised texture segmentation. A modified version of a Siamese network \cite{Siamese2,FullySiamese} was used in \cite{Siamese} 
to segment textures. Fully Convolutional Network (FCN) \cite{FCN} are powerful architectures for pixel semantic 
segmentation and inspired a specific algorithm, called FCNT \cite{FCNT}, to segment textured images. This approach corresponds to an optimized architecture which discards very 
deep shape 
information and makes use of the very shallow features. However, this approach performs poorly on grayscale datasets and needs a refinement procedure to improve its 
performance. It signifies that this architecture mostly relies on the color information and does not really extract textural information. In \cite{DeepVisual}, the authors 
incorporate multi-scale saliency information in a CNN architecture to predict human eye fixation. If the initial purpose of this approach was quite different, this network can 
also be easily modified to segment textures. The V-Net architecture used in \cite{vnet} for medical image segmentation cannot be directly used for texture segmentation, but its 
underlying U-Net \cite{unet} architecture can be utilized for such purpose.

If most of these networks work perfectly on extracting deep semantic features, they usually perform weakly in extracting low-level textures characteristics, 
especially in the presence of different illumination or similar geometric structures. Therefore, our idea is to combine an FCN architecture with empirical wavelet features. The EWT 
will extract the relevant low-level texture information which will be used to feed an FCNT. Since the empirical wavelet filter bank depends on the considered texture, we propose 
an approach to build a unique set of empirical filters for a given dictionary of textures. To resume, the training set of textures is first used to build an ``optimized'' set of 
empirical wavelet filters, then this filter bank is applied to build the texture descriptors utilized to feed the FCNT in both the training and testing stages.

The remainder of the paper is as follows. In section~\ref{sec:EWT}, we briefly recall the construction of the EWT and present our approach to obtain a unique set of empirical 
filters for a given training dataset. In section~\ref{sec:FCNT}, we recall the FCNT architecture and show how the EWT features are used to feed this architecture. Experimental 
results based on several well-known texture datasets are presented in section~\ref{sec:results}. Conclusions and perspectives are given in section~\ref{sec:conc}.

\section{Texture Features Extraction by EWT}
\label{sec:EWT}
Standard wavelet methods are based on bandpass filters those supports in the Fourier domain follow a prescribed scheme driven by the used scaling law. This corresponds to use a prescribed 
partitioning of the Fourier domain. For instance, the classic dyadic wavelet transform aims at iterating a split in half of the lower frequency subband at each scale 
increase. The main disadvantage is that such prescribed scheme does not guarantee that the harmonic modes within the signal will be supported by this partitioning. This 
can be a real issue for texture analysis since it is well-known that the harmonic information is important to characterize textures. In order to circumvent this problem, in 
\cite{EWT1D}, the author proposes an adaptive wavelet transform which automatically finds the Fourier supports which best separate the harmonic modes present in the analyzed 
signal. In \cite{EWT2D} the empirical wavelet transform was extended to 2D in different ways (tensor approach, Littlewood-Paley wavelets, curvelets) for image 
processing purposes. Following the results obtained in \cite{UTSEWT}, we propose to use the empirical curvelet transform to extract texture features. For the 
reader's convenience, we briefly review the principles of the 1D empirical wavelet transform (EWT) and then focus on the 2D curvelet version.

\subsection{1D Case}
As mentioned previously, the EWT is a data-driven approach whose purpose is to separate harmonic modes by detecting their corresponding supports in the Fourier domain and to build a 
Littlewood-Paley bandpass filter associated with each support. Finally, the input signal is filtered by this EWT filter bank to perform the modes separation. In 1D, it is then 
possible to use the Hilbert transform to extract a very accurate time-frequency representation (see \cite{EWT1D} for more details).
 
\begin{figure}[!t]
\includegraphics[width=1\columnwidth]{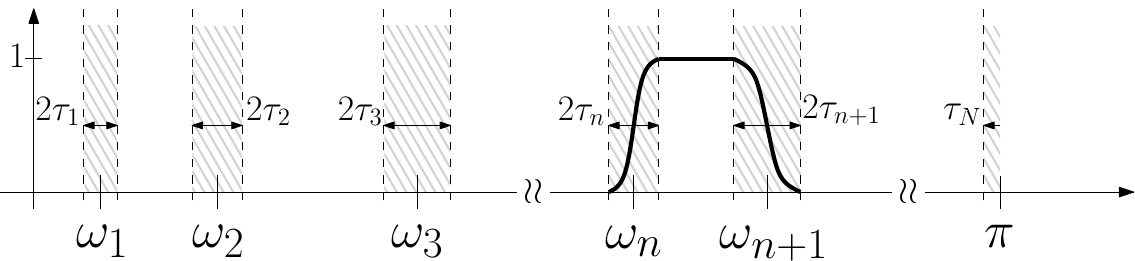}\\
\caption{Magnitude of the Fourier spectrum segmentation and empirical wavelet construction principle.}
\label{fig:EWT1D}
\end{figure}

Let us denote $\omega$ be the normalized frequency (i.e. $\omega\in [0,\pi]$). Let us assume that $N$ supports are detected, we define the set of ordered boundaries (see Fig. 
\ref{fig:EWT1D}) by $(\omega_{n})_{n\in[0,...,N]}\in[0,\pi]$ with the convention that $\omega_{0}= 0$ and $\omega_{N}= \pi$. The shaded areas in 
Figure~\ref{fig:EWT1D}, defined by the 
coefficients $\tau_n$, are needed in the definition of the wavelet filters. It is shown in \cite{EWT1D}, that for a properly chosen $\tau_n$, the corresponding set of 
empirical wavelets forms a tight frame. Several approaches were 
proposed in \cite{EWT1D} and \cite{EWT2D} to detect these boundaries, assuming that $N$ is known. Later, a parameter free approach based on the scale-space theory was proposed in 
\cite{scalespace} also permitting to automatically find $N$. In this paper, we will use this fully automatic method, we refer the reader to \cite{EWT1D} and \cite{scalespace} for more 
details.

\subsection{2D Case}
In \cite{EWT2D}, the authors extended the EWT to
several types of two-dimensional transforms for image processing purposes. In \cite{UTSEWT}, the authors showed that the empirical curvelet transform provides the best texture 
descriptors for 
unsupervised classification. Therefore, we also adopt the empirical curvelet transform (EWT2DC) to extract texture characteristics in this paper. Hereafter, we 
briefly review how the construction of 
empirical curvelet filters is performed, and we refer the reader to \cite{EWT2D} for more
details and examples.

Since scales and orientations are important characteristics in texture analysis, this justifies the effectiveness of the empirical curvelets. Indeed, this transform aims at 
extracting 
harmonic modes supports as polar wedges in the Fourier domain. The major difference with the classic curvelet transform, which also uses a prescribed scheme to partition the 2D 
Fourier plane, is that the EWT2DC aims at detecting the position of these wedges. This corresponds to find a set of radii and angles in the frequency domain. Three options are 
possible to create such partition: 1) finding a set of angles and scales independently of each other (EWT2DC1), 2) finding a set of angles
for each scale ring (EWT2DC2) and 3) finding a set of scale radii for each angular sector (EWT2DC3). The reader is invited to read the papers \cite{EWT1D,EWT2D} and \cite{UTSEWT} 
for complete details on how these partitions are built. Since in \cite{UTSEWT}, the best results were obtained using EWT2DC1, we will only use this option in this paper.

\subsection{Construction of a unique set of EWT2DC filters for a given dictionary of textures}
As described in the previous section, the EWT2DC aims at constructing one specific filter bank for one image. Since in this paper we are investigating supervised classification, 
it means 
that we have a dictionary of textures available to train our algorithm. Therefore, this implies that each texture image from this dictionary will have its own set of curvelet 
filters. However, since we want to characterize all textures in that dictionary, we need to find a unique set of empirical curvelet filters which will be used for all textures. 
Hereafter we propose a general method to merge several sets of boundaries detected on distinct spectra (or histograms) in order to form the expected unique set of 
boundaries. 

\subsubsection{Merging boundary sets}
\begin{figure*}[!t]
\begin{tabular}{cccccc}
\includegraphics[width=0.3\columnwidth]{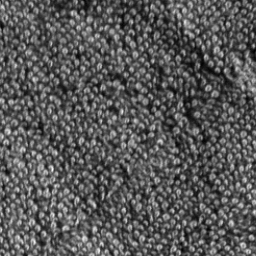} &
\includegraphics[width=0.3\columnwidth]{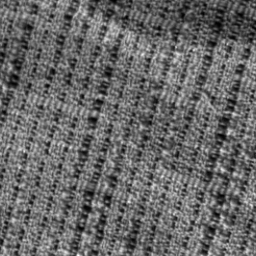} & \includegraphics[width=0.3\columnwidth]{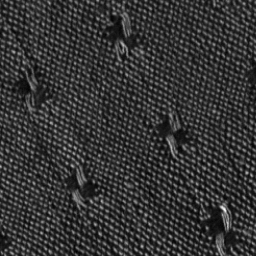} &
\includegraphics[width=0.3\columnwidth]{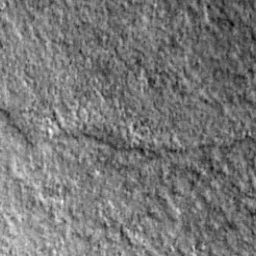} & \includegraphics[width=0.3\columnwidth]{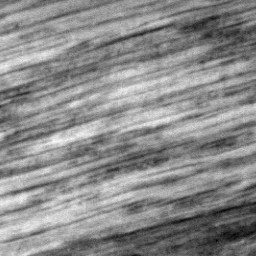} &
\includegraphics[width=0.3\columnwidth]{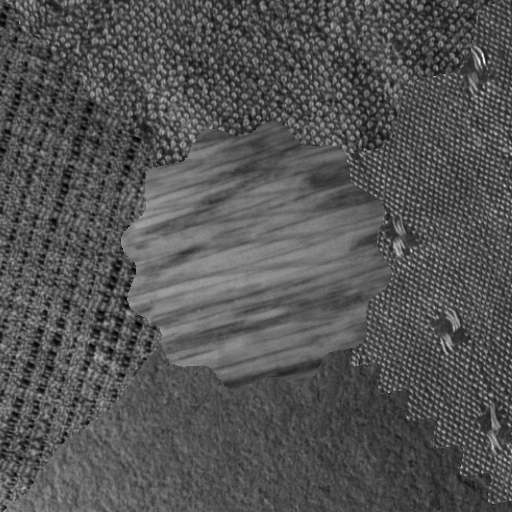}  \\
\\
\includegraphics[width=0.3\columnwidth]{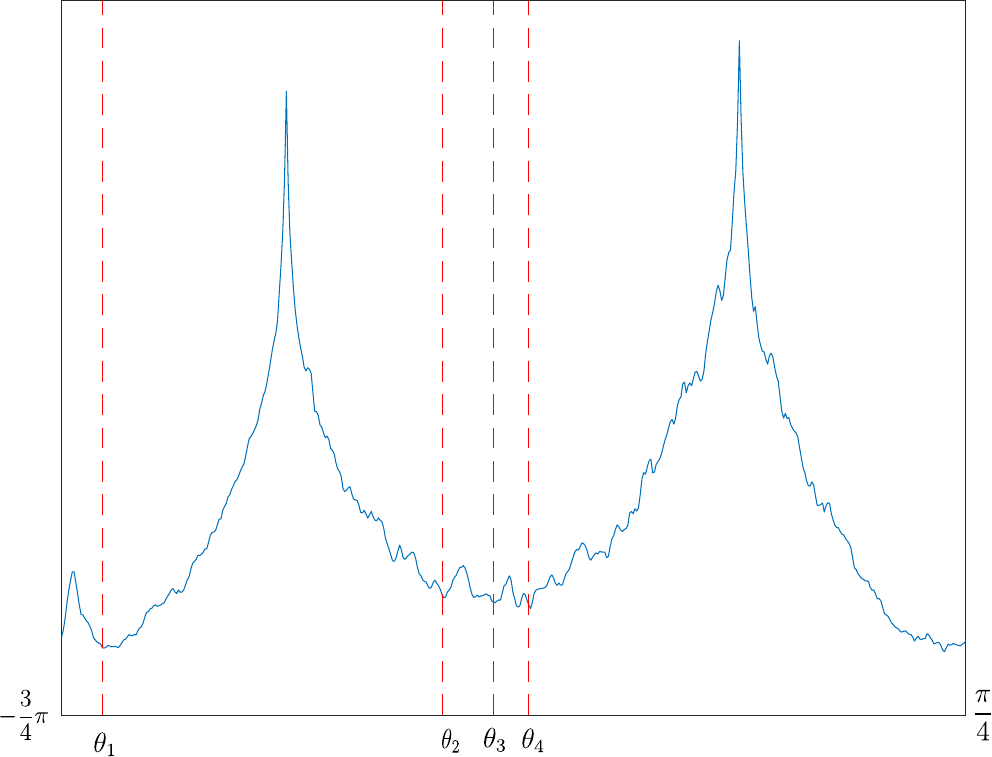} &
\includegraphics[width=0.3\columnwidth]{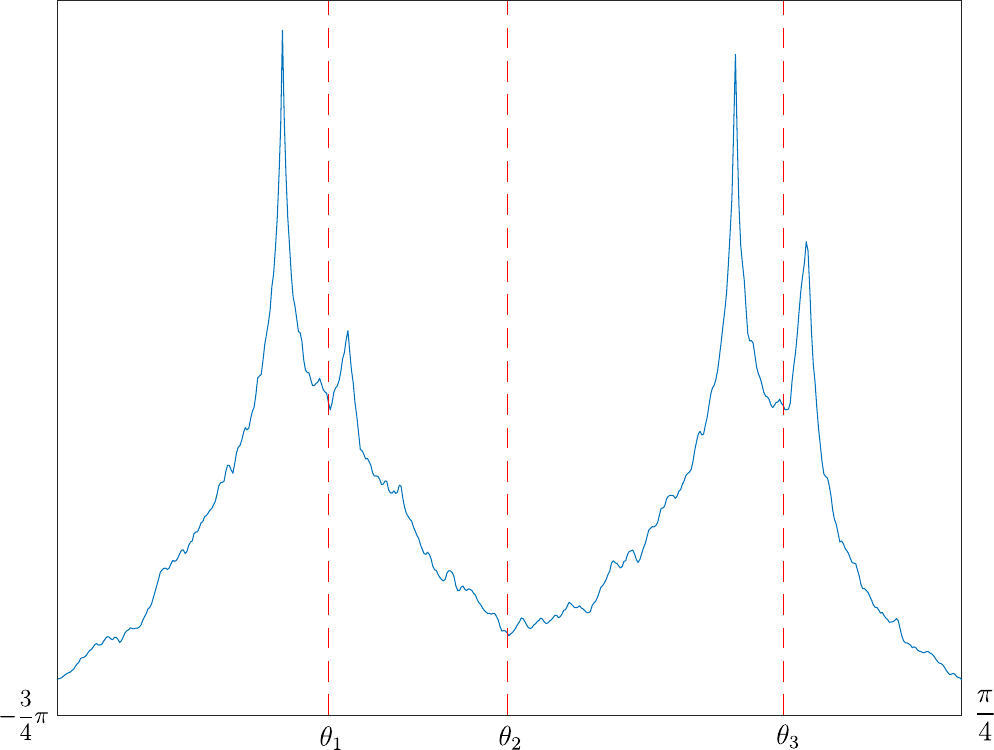} & \includegraphics[width=0.3\columnwidth]{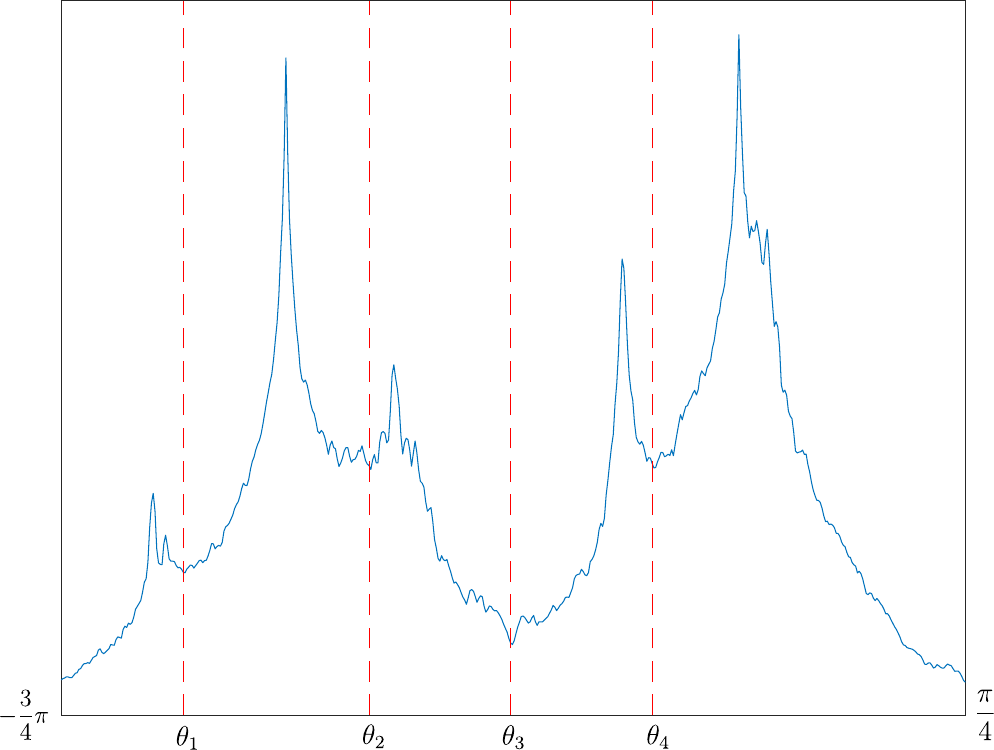} &
\includegraphics[width=0.3\columnwidth]{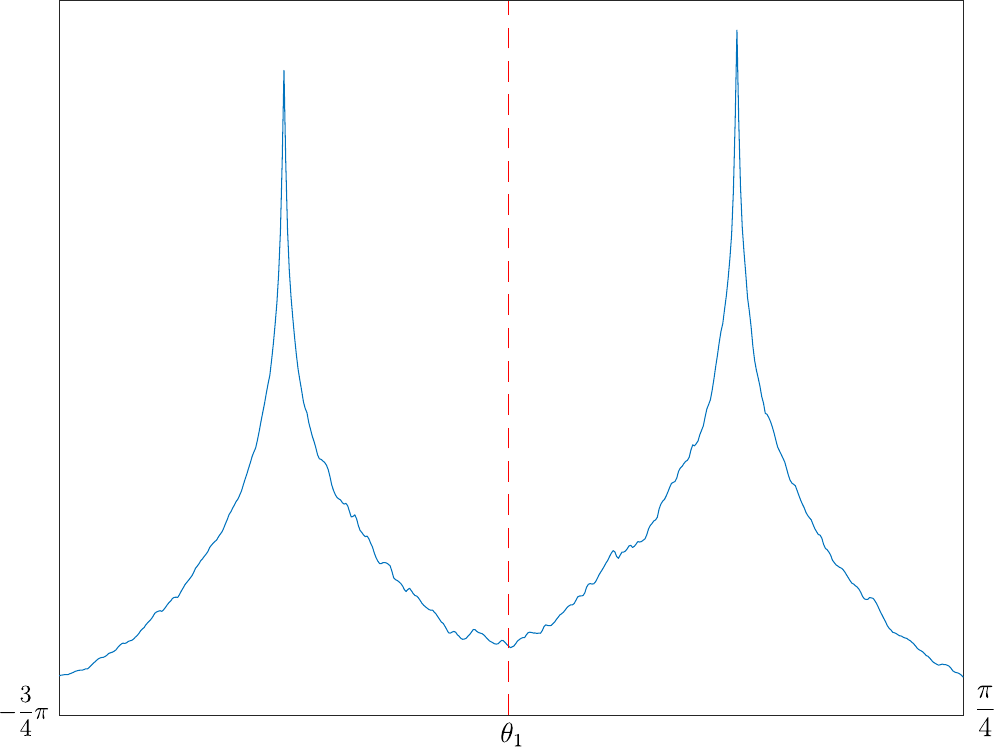} & \includegraphics[width=0.3\columnwidth]{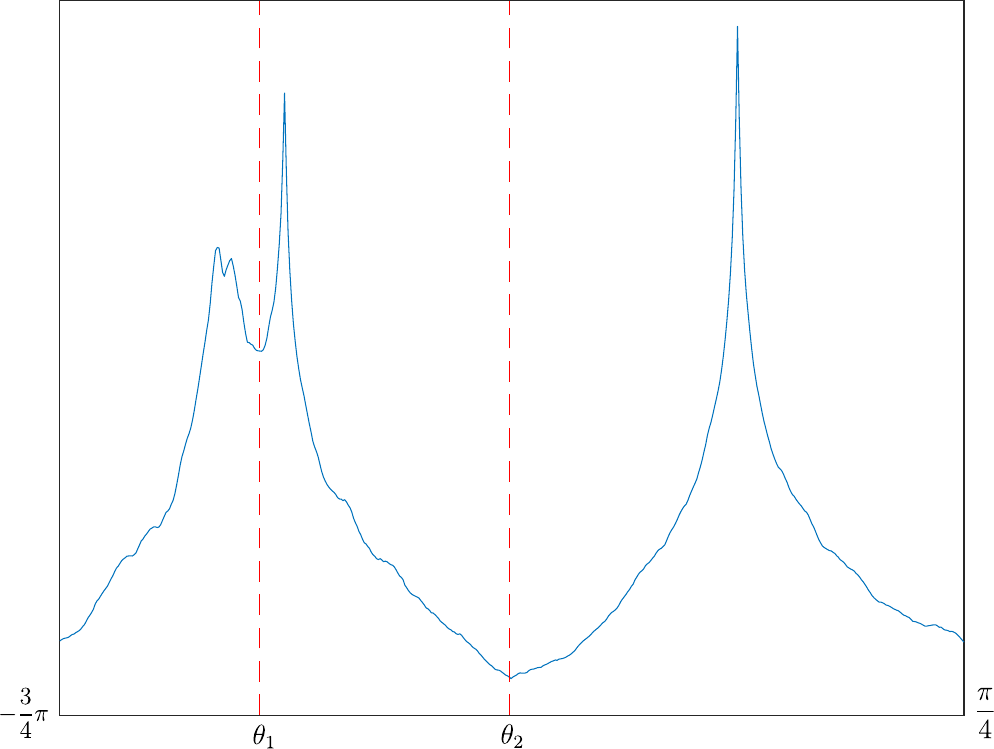} &
\includegraphics[width=0.3\columnwidth]{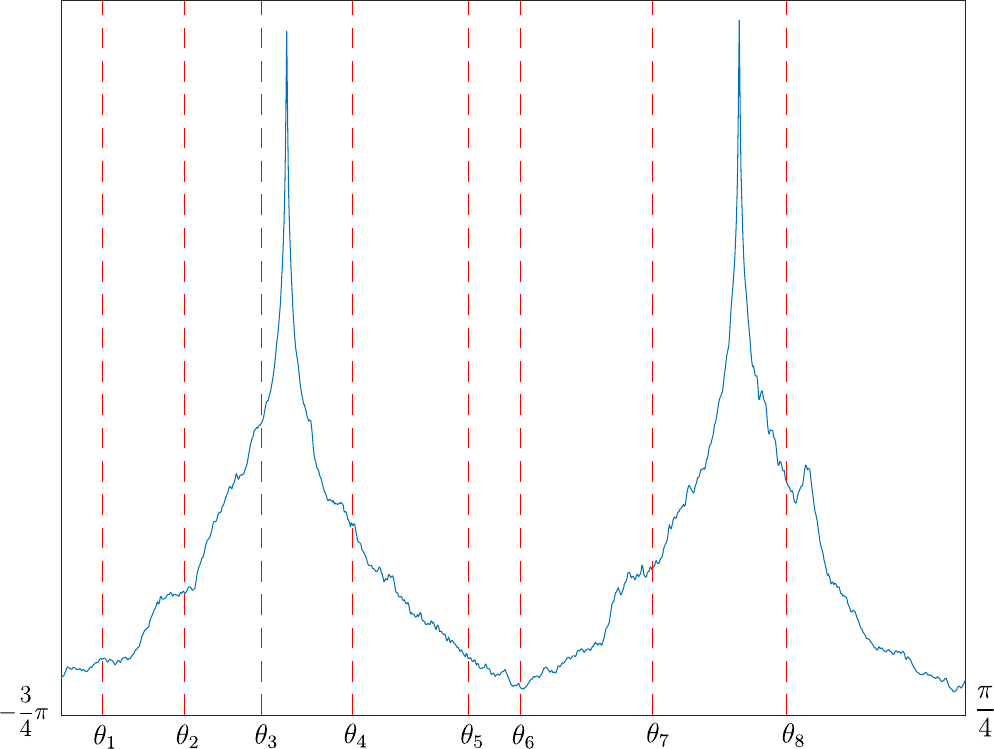} \\
\end{tabular}
\caption{Examples of angular boundaries, i.e. the sets $\Omega_{\theta}^{i}$ (the vertical red lines), obtained from different input textured images.}
\label{fig:boundary}
\end{figure*}

Let us assume we have $N_T$ 1D spectra, denoted $\mathcal{F}_i(\omega),i=1,\ldots,N_T$. Applying the scale-space boundary detection method described in \cite{scalespace} to each 
spectrum, 
we get $N_T$ sets of boundaries that we will denote $\Omega^i=\{\omega_n^i\}_{n=0,\ldots,N^i}$ where we keep the convention that $\omega_0^i=0$ and $\omega_{N^i}^i=\pi$. We first 
merge these sets by taking their union to get the set 

\begin{equation}
\Omega=\bigcup_{i=0}^{N_T}\Omega^i=\{\omega_n\}_{n=0,\ldots,N},
\end{equation}
where $N$ is the number of boundaries minus one present in $\Omega$ (we also assume $\Omega$ is a well-ordered set, i.e. $\forall n=0,\ldots, N-1,\omega_{n+1}>\omega_n$). Such 
merging does not guarantee that each corresponding support $[\omega_n,\omega_{n+1}]$ corresponds to a harmonic mode in any of the original spectra and thus should be removed. We 
propose a two stages 
method to remove these useless supports. The first stage is based on the principle that if a support corresponds to a harmonic mode, it must contain one of the highest local 
maxima present in the supports of the original spectra $\mathcal{F}_i$. Let us denote 
$\Lambda^i=\{\lambda_n^i\}_{n=0,N^i-1}$ the set containing the positions of the largest local maximum on each support generated by $\Omega^i$, i.e. $\forall i=1,\ldots,N_T-1,\forall 
n=0,\ldots,N^i$,
\begin{equation}
\lambda_n^i=\underset{\omega\in [\omega_n^i,\omega_{n+1}^i]}{\arg\max}|\mathcal{F}_i(\omega)|.
\end{equation}
Then, a support $[\omega_n,\omega_{n+1}]$ defined from $\Omega$ contains a harmonic mode if at least one position $\lambda_n^i$ belongs to that support, i.e. $\exists 
i,k,\lambda_k^i\in[\omega_n,\omega_{n+1}]$. If such $\lambda_k^i$ exists then we keep these boundaries, otherwise we replace them by their midpoint, i.e. 
$(\omega_n+\omega_{n+1})/2$. Thus, we obtain an updated set of boundaries $\tilde{\Omega}=\{\tilde{\omega}_n\}_{n=0,\ldots,\tilde{N}}$. This process is repeated until all remaining supports contains at least one of the initial local maxima $\lambda_n^i$.

The second stage consists in removing the ``smallest'' supports as we expect that each harmonic mode contains enough useful information. To perform this step, we simply compute 
the width of each support $\tilde{\omega}_{n+1}-\tilde{\omega}_n$ and if this quantity is larger than a threshold $T$ (the choice of $T$ depends on the processed histogram and will be discussed in the next section), we keep these boundaries, otherwise we 
merge them using again the midpoint rule described above. This process is repeated until all supports are of width larger than $T$. This finally gives us our final set of 
boundaries $\Omega^F=\{\omega_n\}_{n=0,\ldots,N^F}$. The 
whole merging algorithm is resumed in Algorithm~\ref{alg:merging}.

\begin{algorithm}[htb]
\caption{Boundary sets merging algorithm.}
\label{alg:merging}
\begin{algorithmic}[1]
\Require $N_T$ spectra $\mathcal{F}_i$.
\State Apply the scale-space method on each spectrum to obtain their corresponding set of boundaries $\Omega^i$.
\State Find the sets $\Lambda^i$ of local maxima on each support.
\State Merge all sets of boundaries to obtain $\Omega$.
\State Remove all supports which do not contain any of the local maximum to get the set $\tilde{\Omega}$.
\State Remove all supports of length less than the threshold $T$ to obtain the final set of boundaries $\Omega^F$.
\end{algorithmic}
\end{algorithm}

\subsubsection{Definition of the empirical curvelet filters}
\begin{figure*}[!t]
\begin{center}
\includegraphics[width=2\columnwidth]{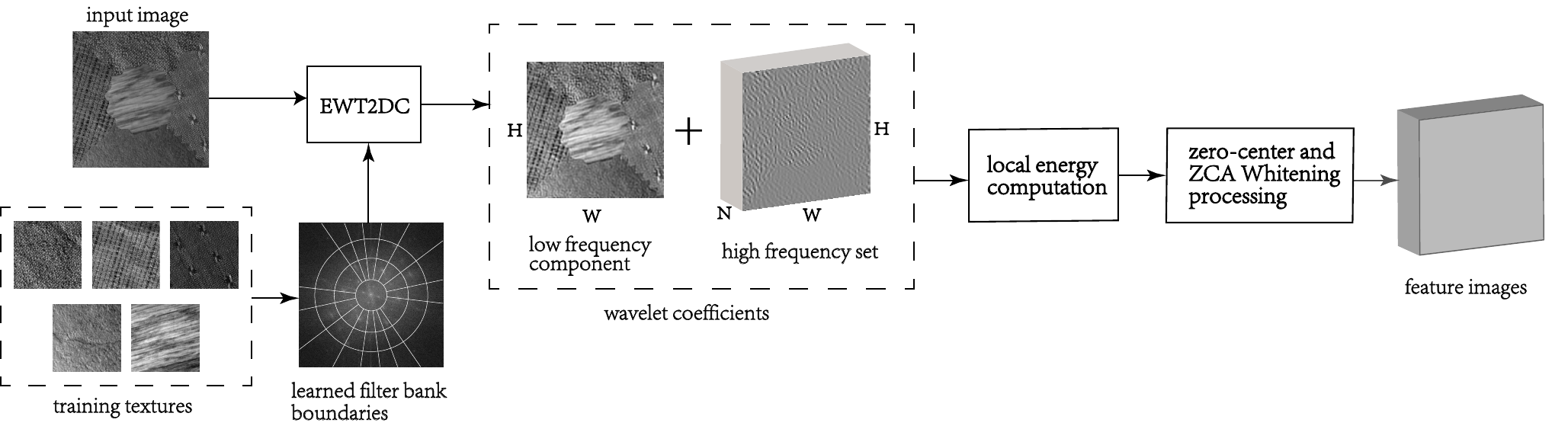}\\
\caption{Proposed texture feature extraction process.}
\label{fig:ewtfeature}
\end{center}
\end{figure*}

The construction of the empirical curvelet filters is based on the procedure described in \cite{EWT2D}. For each input image $i$, we detect a set of scale boundaries, denoted 
$\Omega_{\omega}^{i}=\{\omega^{n}_{i}\}_{n=0,...,n=N_s^i}$, and a set of angular boundaries, denoted $\Omega_{\theta}^{i}=\{\theta_{i}^{m}\}_{m=1,...,N_{\theta}^i}$, where 
$N_{s}^i$ and $N_{\theta}^i$ are respectively the number of detected scales and angular sectors for the image $i$. Figure~\ref{fig:boundary} illustrates the sets $\Omega_{\theta}^{i}$ (the vertical red lines) corresponding to several input textured images.
Then we use the merging technique described above on 
$\{\Omega_{\omega}^{i}\}_{i=1,\ldots,N_T}$ and $\{\Omega_{\theta}^{i}\}_{i=1,\ldots,N_T}$ to remove the useless supports (the thresholds used to detect ``small'' supports are fixed 
to $T_\omega=0.2$ and $T_\theta=0.07$, respectively). This process provides us two final sets of scales and angular boundaries denoted 
$\Omega_{\omega}=\{\omega_n\}_{n=0,\ldots,N_s}$ and $\Omega_{\theta}=\{\theta_m\}_{m=0,\ldots,N_\theta}$, respectively. 

Using the above notations, the empirical curvelet filters are then made of: a lowpass filter ($\phi_1$) defined by ($\omega$ and $\theta$ correspond to the polar coordinates in the 
Fourier domain and $\mathcal{F}_2$ the 2D Fourier transform), $\forall\theta\in[0,2\pi]$,
\begin{equation}
\mathcal{F}_2(\phi_{1})(\omega,\theta)=
\left\{
\begin{array}{lr}
  				1 \qquad\text{if $|\omega|\le(1-\gamma)\omega_{1}$,} \\  
				\cos\left[\frac{\pi}{2}\beta\left(\frac{1}{2\gamma\omega_{1}}\left(|\omega|-(1-\gamma)\omega_{1}\right)\right)\right] \\
				 \qquad\text{if $(1-\gamma)\omega_{1}\le|\omega|\le(1+\gamma)\omega_{1}$,} \\
				 0 \qquad\text{otherwise,} \\
\end{array}
\right.
\label{psi1}
\end{equation}
The function $\beta$ is an arbitrary function belonging to $\mathcal{C}^k([0,1])$, fulfilling the properties $\beta(x)=0$ if $ x\le0$, $\beta(x)=1$ if $x\ge1$ and 
$\beta(x)+\beta(1-x)=1,\forall x\in[0,1]$. A classic choice is given by $\beta(x)=x^4(35-84x+70x^2-20x^3)$. The parameter $\gamma$ allows us to ensure that only two consecutive 
filters can overlap. We refer the reader to \cite{EWT2D} for more details about these two parameters and how they can be automatically chosen. Like for standard 
curvelets, the bandpass curvelet filters ($\psi_{nm}$) correspond to polar wedges in the Fourier domain and are defined by the product of a radial window $W_n$ and a polar window 
$V_m$, i.e.
\begin{equation}
\mathcal{F}_2(\psi_{nm}) = W_{n}V_{m}
\end{equation}
where $W_{n}$, if $n\neq N_s-1$, is given as:
\begin{equation}
W_{n}(\omega)=
\left\{
\begin{array}{lr}  
             1 \qquad\text{if $(1+\gamma)\omega_{n}\le|\omega|\le(1-\gamma)\omega_{n+1}$,} \\  
             \cos\left[\frac{\pi}{2}\beta\left(\frac{1}{2\gamma\omega_{n+1}}\left(|\omega|-(1-\gamma)\omega_{n+1}\right)\right)\right] \\
             \qquad\text{if $(1-\gamma)\omega_{n+1}\le|\omega|\le(1+\gamma)\omega_{n+1}$,} \\
             \sin\left[\frac{\pi}{2}\beta\left(\frac{1}{2\gamma\omega_{n}}\left(|\omega|-(1-\gamma)\omega_{n}\right)\right)\right] \\
  	    \qquad\text{if $(1-\gamma)\omega_{n}\le|\omega|\le(1+\gamma)\omega_{n}$,} \\
	    0 \qquad\text{otherwise,} \\
             \end{array}  
\right.
\end{equation}
and if $n = N_s-1$,
\begin{equation}
W_{N^{s}-1}(\omega)=
\left\{
\begin{array}{lr}  
             1 \qquad\text{if $(1+\gamma)\omega_{N_s-1}\le|\omega|$,} \\  
             \sin\left[\frac{\pi}{2}\beta\left(\frac{1}{2\gamma\omega_{N_s-1}}\left(|\omega|-(1-\gamma)\omega_{N_s-1}\right)\right)\right] \\
  	    \quad\text{if $(1-\gamma)\omega_{N_s-1}\le|\omega|\le(1+\gamma)\omega_{N_s-1}$,} \\
	    0 \qquad\text{otherwise,} \\
             \end{array}  
\right.
\end{equation}
and $V_m$ is defined as (assuming $2\pi$ periodicity of the angular sectors):
\begin{equation}
V_{m}(\theta)=
\left\{
\begin{array}{lr}  
             1 \qquad\text{if $\theta_{m}+\Delta\theta\le\theta\le\theta_{m+1}-\Delta\theta$,} \\  
             \cos\left[\frac{\pi}{2}\beta\left(\frac{1}{2\Delta\theta}\left(\theta-\theta_{m+1}+\Delta\theta\right)\right)\right] \\
             \qquad\text{if $\theta_{m+1}-\Delta\theta\le\theta\le\theta_{m+1}+\Delta\theta$,} \\ 
             \sin\left[\frac{\pi}{2}\beta\left(\frac{1}{2\Delta\theta}\left(\theta-\theta_{m}+\Delta\theta\right)\right)\right] \\
  	    \qquad\text{if $\theta_{m}-\Delta\theta\le\theta\le\theta_{m}-\Delta\theta$,} \\ 
	    0 \qquad\text{otherwise.} \\
             \end{array}  
\right.
\label{vm}
\end{equation}
Finally, the corresponding filter bank is given by ($\boldsymbol{x}$ denotes a pixel location in the 2D plane)
\begin{equation}
\mathcal{B}^{\mathcal{EC}}=\left\{\phi_{1}(\boldsymbol{x}),\left\{\psi_{nm}(\boldsymbol{x})\right\}_{n=1...N_s-1,{m=1...N_{\theta}}}\right\},
\end{equation}
and we will denote $\mathcal{W}_f^{\mathcal{EC}}(\boldsymbol{x})$ the empirical curvelet transform of an image $f$, i.e.
\begin{eqnarray}\label{eq:ewtc}
\mathcal{W}_f^{\mathcal{EC}}(\boldsymbol{x})=\left((\phi_1\ast f)(\boldsymbol{x}),(\psi_{11}\ast f)(\boldsymbol{x}),(\psi_{12}\ast f)(\boldsymbol{x}),\right.\\ \notag
\left.\ldots,(\psi_{nm}\ast f)(\boldsymbol{x}),\ldots,(\psi_{N_sN_\theta}\ast f)(\boldsymbol{x})\right),
\end{eqnarray}
where $\ast$ denotes the convolution product. In practice, since all filters are defined in the Fourier domain, we perform all convolutions as pointwise products in the Fourier domain and take the inverse Fourier transform to get the curvelet coefficients.

\subsection{Texture feature extraction}
Our main proposal is to utilize empirical curvelets to extract texture features to feed the segmentation network instead of using the original image pixels directly. It is widely 
accepted that wavelet (or curvelet) coefficients need to be post-processed to obtain the expected features. 

Based on \eqref{eq:ewtc}, we denote the curvelet coefficients as $\mathcal{W}_f^{\mathcal{EC}}(\boldsymbol{x}):D\to\mathbb{R}^K$ where $D$ is the image domain and $K$ is the total 
number of curvelet filters. Let us denote $\mathcal{T}^{\mathcal{EC}}(\boldsymbol{x}):D\to\mathbb{R}^K$ the final feature vector at a given pixel $\boldsymbol{x}\in D$.  Its 
$k-$th coordinate 
$\mathcal{T}^{\mathcal{EC}}_{k}(\boldsymbol{x}),k=1,2,...,K$, is obtained by applying a post-processing operator $E$ to the curvelet coefficients, i.e. 
$\mathcal{T}^{\mathcal{EC}}_{k}(\boldsymbol{x})= E[\mathcal{W}^{\mathcal{EC}}_{k}](\boldsymbol{x})$. In the field of wavelet-based feature vectors, local energy is widely used as 
such operator, i.e. 
\begin{equation}
\mathcal{T}^{\mathcal{EC}}_{k}(\boldsymbol{x})=E[\mathcal{W}^{\mathcal{EC}}_{k}](\boldsymbol{x})=\sum_{\boldsymbol{y}\in 
D_{\boldsymbol{x}}}\left|\mathcal{W}^{\mathcal{EC}}_{k}(\boldsymbol{y})\right|^2
\end{equation}
where $D_{\boldsymbol{x}}$ is a window of size $s\times s$ pixels centered at $\boldsymbol{x}$.

It has been shown in \cite{Whitening} that performing a zero-centering and a ZCA (zero-phase component analysis) whitening \cite{ZCA} on the input data improves the performance of convolutional 
neural networks. Let us denote $\mathbf{X}$ the $N_p\times K$ 
feature matrix ($N_p$ is the total number of pixels of the input image) whose rows are the vectors $\mathcal{T}^{\mathcal{EC}}(\boldsymbol{x})$. The output matrix after 
performing the ZCA whitening is obtained by
$\mathbf{Y}=\sqrt{(N_p-1)}\mathbf{P}\mathbf{D}^{-1/2}\mathbf{P}^{\top}\mathbf{X}$, where $\mathbf{X}\mathbf{X}^{\top}=\mathbf{P}\mathbf{D}\mathbf{P}^{\top}$, $\mathbf{P}$ is an 
orthogonal matrix and $\mathbf{D}$ a diagonal matrix. The updated features 
in $\mathbf{Y}$ have less correlation and hence are more discriminating. This matrix $\mathbf{Y}$ will be used to feed the deep learning network described in the next section. The 
whole process corresponding to texture features extraction is illustrated in Fig.~\ref{fig:ewtfeature}. 

\section{Fully Convolutional Network based Segmentation}\label{sec:FCNT}
In this section, we present the overall architecture used in the proposed network. Since our architecture is based on fully convolutional networks (FCN), we start to give a quick 
review of FCN. Next we present the network architecture proposed to segment grayscale 
texture images. Finally, we extend this architecture to process color texture images.

\subsection{Fully Convolutional Networks}

\begin{figure}[!t]
\begin{center}
\includegraphics[width=0.9\columnwidth]{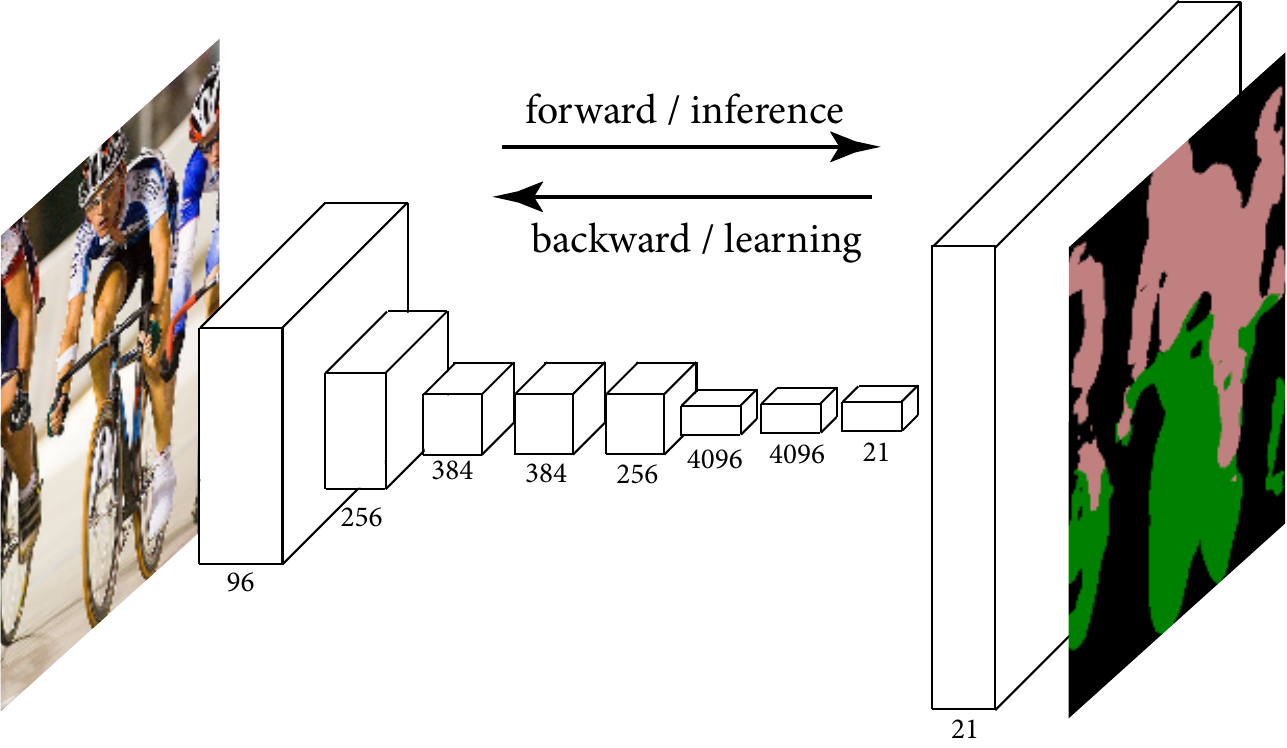}\\
\caption{Standard fully convolutional network makes dense
predictions for per-pixel semantic segmentation}
\label{fig:fcn}
\end{center}
\end{figure}
Fully Convolutional Networks (FCN) were proposed in \cite{FCN} to perform semantic segmentation of images and have shown their superiority among all other deep learning networks. The 
idea is to adapt existing network architectures to perform a dense classification pixelwise. The first layer is the image itself and the other layers are all convolutional 
networks which include three basic components: convolution, pooling and activation functions. These components work on local input regions and only depend on corresponding 
spatial coordinates. An advantage of an FCN is that they naturally work on input images of arbitrary size and provides an output with the same size. The learning and 
inference stages are performed on the whole image at a time instead of processing patches independently of each other. This is made possible through dense feedforward 
calculation and back propagation processing. The back propagation step performs an interpolation of the coarse outputs by using an upsampling procedure whose weights can also 
eventually be 
learned. This step can be seen as a dense pixel backward convolution. Fig.~\ref{fig:fcn} presents an instance of FCN with an input image and its output pixel 
prediction. In \cite{FCN}, the author also proposed a skip architecture to take advantage of the features in different layers which combines deep, coarse information with shallow 
and 
fine information to further refine the spatial precision of the output. These different architectures are called FCN-32s, FCN-16s and FCN-8s, depending on which levels are skipped,
respectively.

\subsection{Network architecture to segment grayscale images}

\begin{figure*}[!t]
\includegraphics[width=2\columnwidth]{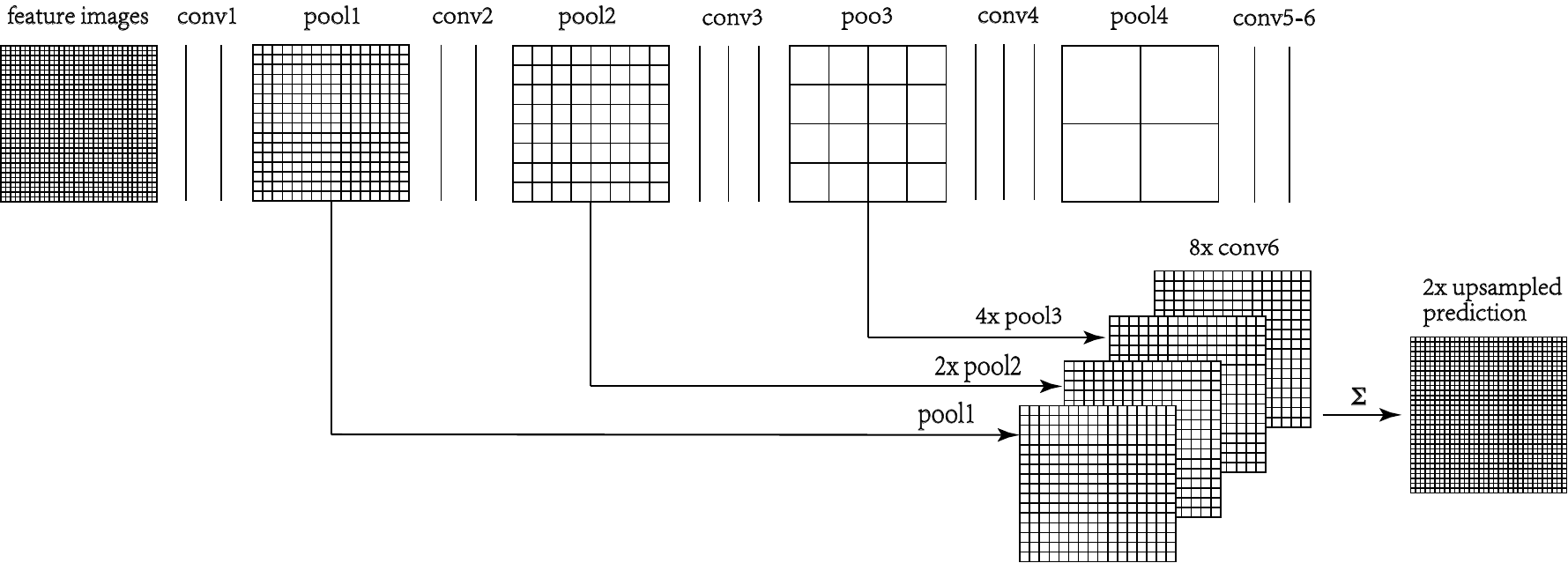}\\
\caption{Network architecture used to segment grayscale images.}
\label{fig:gfcn}
\end{figure*}
Recently, in \cite{FCNT}, the authors proposed a modified FCN architecture, called FCNT, to specifically address the problem of texture segmentation. The FCNT is a modified 
version of FCN-8s having one less convolutional layer based on the idea that the complexity of texture patterns is less than the shape complexity of semantic segmentation (see
\cite{FCN} for more details). Moreover, the skip connections in FCNT are also shifted to the shallower layers of the network, i.e. the first, second and third convolutional 
blocks. In this paper, for grayscale texture images, we first use the FCNT architecture but we feed the network with the empirical wavelet features rather than the original pixel values. Our architecture based on FCNT is depicted in Fig.~\ref{fig:gfcn} which also provides an illustration of the skip architecture mentioned 
earlier. 
Furthermore, we want to emphasize the reader that the same idea of using the empirical wavelets to feed more complex architectures such as U-Net 
\cite{unet}, Siamese-Net \cite{Siamese}, Deep Visual Attention Model (DA) \cite{DeepVisual} and PSP-Net \cite{pyramid} applies and will be used in our experiments 
to assess the effectiveness of our proposed architecture.

\subsection{Network architecture to segment color images}

\begin{figure*}[!t]
\includegraphics[width=2\columnwidth]{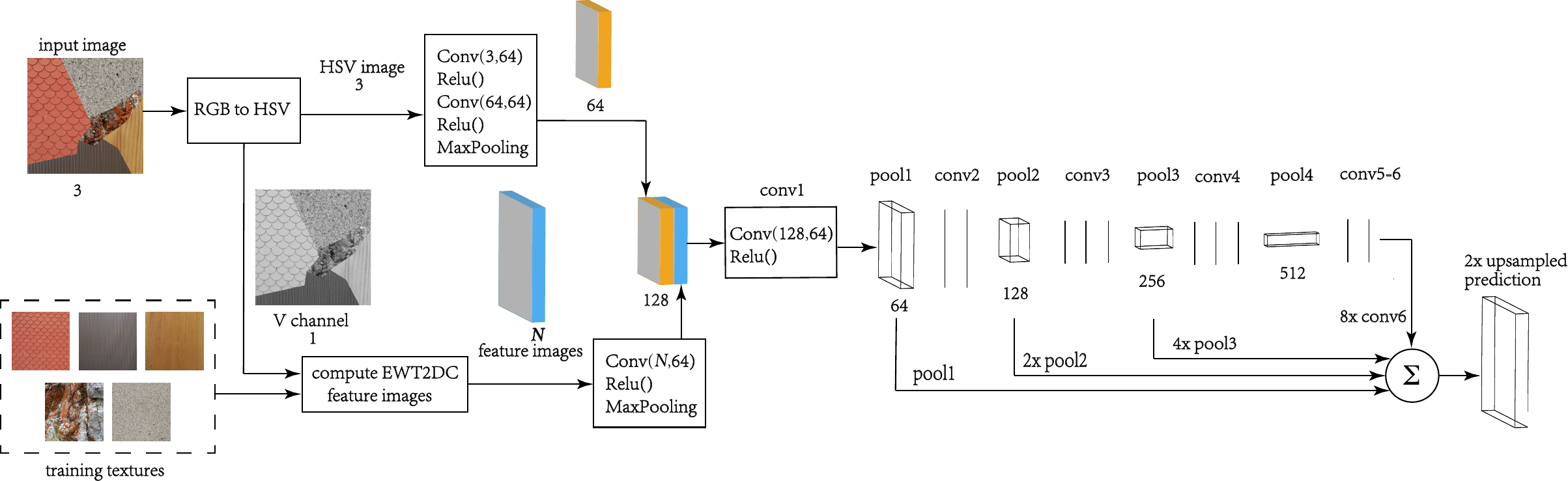}\\
\caption{Network architecture used to segment color images }
\label{fig:cfcn}
\end{figure*}

To segment color images, we propose an architecture which combines both the empirical curvelet features and the original colors through the HSV channels of the input image. 
Assuming that all input images are in the HSV format, we use the V channel to extract the texture feature information, i.e. the EWT2DC filter bank is constructed by analyzing all V 
channels from each texture image from the dictionary. When testing a new image, the empirical curvelet features are then extracted from the same V channel by the 
previously obtained empirical curvelet filter bank. The HSV channels are used to feed the first two convolutional layers in order 
to extract the color information while the texture features (extracted from the V channel only) are used to feed another convolutional layer to extract the geometry of the texture 
information. Then, these two groups of layers (128 channels) are used to feed a simplified version of FCNT where the prediction layers are obtained by summing the information from 
the intermediate pool layers. The whole architecture is illustrated in Fig.~\ref{fig:cfcn}. Like for the grayscale case, the EWT features can be combined with the 
color information with more complex networks and will be assessed in the experiment section.

\section{Experiment Results}\label{sec:results}
In this section, we present the obtained results when applying the architectures described in the previous sections. We implement the proposed empirical wavelet features 
extraction 
method described in Section \ref{sec:EWT} in Matlab by using the functions available in the Empirical Wavelet Transform toolbox\footnote{Available at 
\url{https://ww2.mathworks.cn/matlabcentral/fileexchange/42141-empirical-wavelet-transforms}}. In addition, we utilize the PyTorch toolbox to implement all different networks on a single GTX 
1080Ti external GPU with 11GB memory. To train the segmentation network, we use the Adam optimizer where the learning rate is fixed to $10^{-3}$, the decay weight to $10^{-4}$, 
$\beta_1$ is set to $0.95$ and $\beta_2$ to $0.999$. All experiments were conducted on a laptop computer with 2.5GHz quad-core Intel Core i7 processor and 16GB memory. 

We test our approaches on three different datasets and compare the segmentation results with existing algorithms. Grayscale experiments are conducted on the Outex 
\cite{Outex} and UIUC \cite{UIUC} datasets while the Prague dataset \cite{Prague} made of mosaics of colored textures is used to test the color based algorithm. Qualitative and 
quantitative 
results are provided in the following sections for assessment. 

\subsection{Grayscale textures}
\begin{figure}[!t]
\includegraphics[width=1\columnwidth]{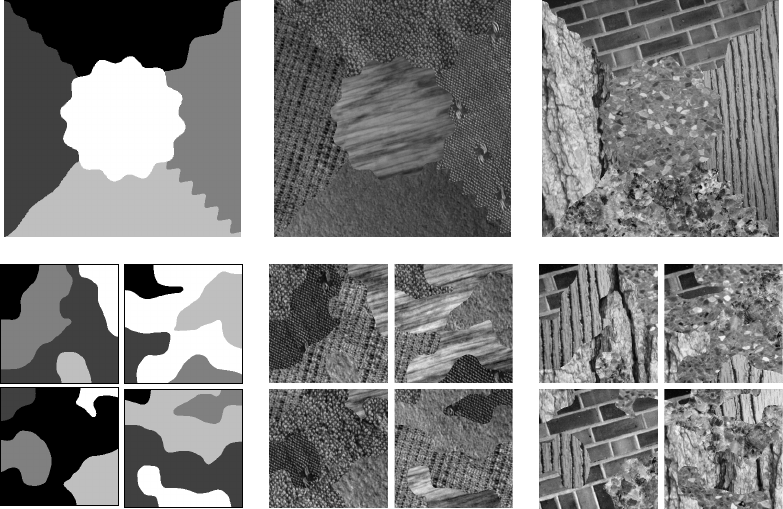}\\
\caption{The used ground-truths for test images and training images (left) and some corresponding test image and training samples in Outex dataset (center) and UIUC 
dataset (right).}
\label{fig:gray_gt}
\end{figure}

To assess the performance of the proposed algorithm for grayscale textures, we select two popular texture datasets: Outex \cite{Outex} and UIUC \cite{UIUC}. The Outex 
dataset has 100 composite texture images, each one being a composite of five training textures. These are generated by mixing twelve different texture images with 
different rotations according to the regions depicted by the ground-truth shown in the top-left row of Fig.~\ref{fig:gray_gt}. The UIUC dataset does not contain as many test 
images as the Outex dataset, therefore we decided to build similar sets of test images as the Outex dataset following the same procedure: we selected thirteen pristine textures 
and randomly composed a hundred test images using the same ground-truth images given on the top-left row of Fig.~\ref{fig:gray_gt}. The size of all test images is $512\times512$ 
pixels and the size of all pristine textures for training is $256\times256$. Moreover, all images are encoded on 256 grayscales.

Regarding the training stage, both datasets only have five different training pristine textures. Unfortunately, only five training textures are not 
sufficient to efficiently learn the different weights involved in the deep neural networks. To solve this issue, we generated an augmented training set of a 1000 images (for each 
dataset) from five 
pristine textures by randomly generating ground-truths images. These ground-truths are built via the following process: 1) we create a $256\times 256$ random image following a 
normal distribution within the range $[0,1]$, 2) we apply a Gaussian filtering ($\sigma = 10$), 3) we binarize this filtered image using a threshold corresponding to the median value of the image. Then each connected components in that binary image are labeled to provide different regions (since we only have five pristine textures, we only keep the generated masks which provide five regions). Finally, each pristine texture is assigned to each mask region to obtain a training image. 
The second row of Fig.~\ref{fig:gray_gt} presents several created mask and their corresponding training mosaics. 

The window size used in the extraction of the texture features is set to 1 (i.e. there is no spatial averaging) as we experimentally found that it is the value which gives the best 
results. Moreover, since the textures in these datasets mainly contain high frequency components, we simply discard the use of the low frequency component ($\phi_1\ast f$) from the empirical 
curvelet coefficients.

Several approaches were proposed in the literature to quantitatively assess segmentation performances. These methods are mainly based on different 
interpretations of the notion of partition. In this paper, we use the following metrics: normalized variation 
of information (NVOI), swapped directional hamming distance (SDHD), Van Dongen distance (VD), swapped segmentation covering (SSC), bipartite graph matching (BGM) and bidirectional 
consistency error (BCE). The reader is referred to \cite{benchmark} for more details and interpretations of these metrics. All these metrics provide a number between 0\% and 
100\%, the latter corresponding to perfect segmentation. The computation of these metrics is made easy via the availability of the SEISM Toolbox\footnote{\url{https://github.com/jponttuset/seism}}.

\begin{figure}[!t]
\includegraphics[width=0.9\columnwidth]{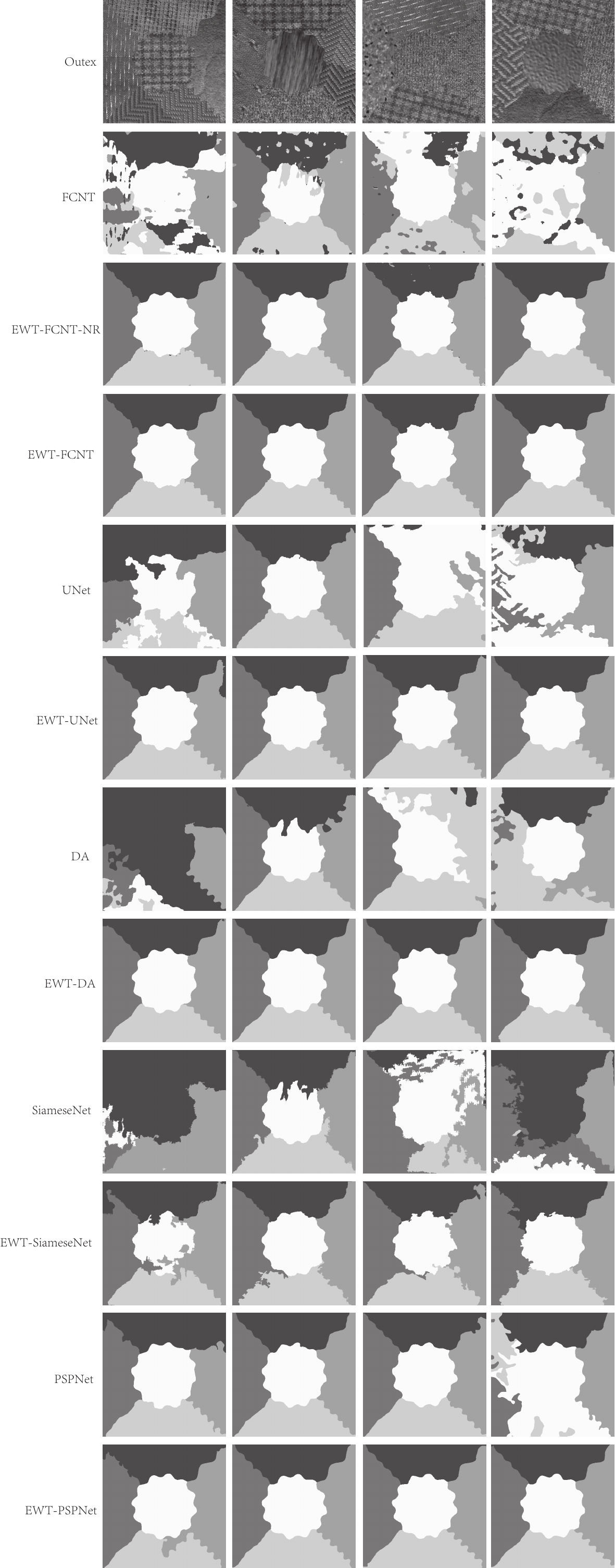}
\caption{Grayscale experiment results on the Outex dataset. First row: input test image to segment. Second row: FCNT segmentation without refinement. Third and fourth row: segmentation from our proposed 
approach without and with refinement, respectively. The fifth to twelve row illustrate the results corresponding to the original and empirical curvelet based approaches for the U-Net, DA, Siamese-Net and PSP-Net architectures.}
\label{fig:outex_result}
\end{figure}

\begin{figure}[!t]
\includegraphics[width=0.9\columnwidth]{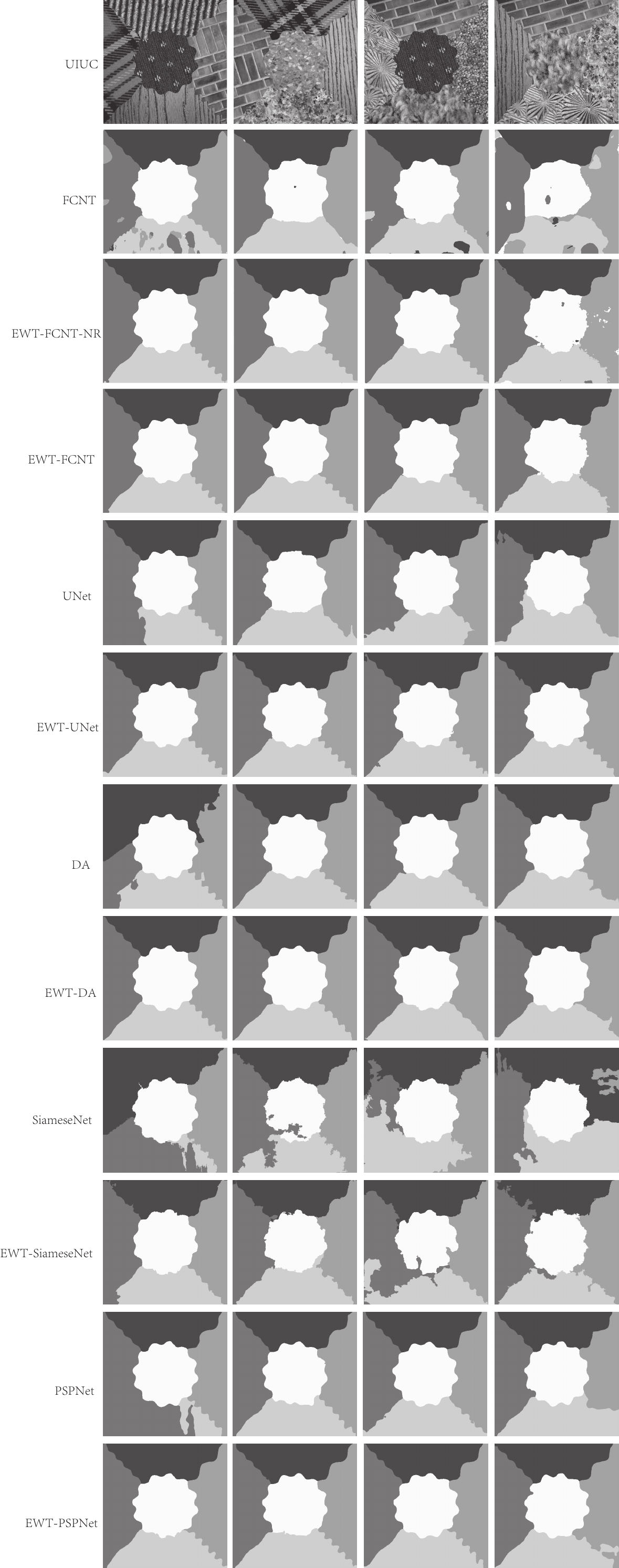}
\caption{Grayscale experiment results on the UIUC dataset. First row: input test image to segment. Second row: FCNT segmentation without refinement. Third and fourth row: segmentation from our proposed 
approach without and with refinement, respectively. The fifth to twelve row illustrate the results corresponding to the original and empirical curvelet based approaches for the U-Net, DA, Siamese-Net and PSP-Net architectures.}
\label{fig:uiuc_result}
\end{figure} 

Fig.~\ref{fig:outex_result} and Fig.\ref{fig:uiuc_result} show some segmentation results obtained on the Outex and UIUC datasets, respectively. The first row of these figures depict  
some test images. The second row show the results obtained by the classic FCNT approach (i.e. it does not use any specific texture feature). The third row show the results 
obtained by the proposed EWT-FCNT algorithm. The fourth row correspond to our algorithm results after a final refinement step which consists in keeping all regions larger 
than 0.5\% of the total number of pixels of the image while all other regions are relabeled with the class index corresponding to the largest adjacent region. All following rows 
are all refined by this same method. To further illustrate the impact of feeding neural networks with empirical curvelet based features v.s. the pristine pixels, in the consecutive rows we provide the output of the U-Net, Deep Visual Model (DA), Siamese-Net, PSP-Net architectures with and without the use of empirical curvelets.
Visual inspections of these experiments illustrate the efficacy of the proposed method. We also observe that the FCNT and other CNN based algorithms generate a lot of misclassification, especially 
in regions having nonuniform luminosity. In contrast, our method handles very well such problem because luminosity changes mostly affect low frequencies and keep high frequencies 
consistent from one image to the other. This means that the use of empirical curvelets provides more stable textures descriptors.
\begin{table*}[ht]
\centering

\begin{threeparttable}
\caption{\label{tab:g1}Benchmark results for the grayscale experiments.}

\begin{tabular}{cccccccccc}
\toprule[0.7pt]
Method & Refinement & NVOI & SSC & SDHD & BGM & VD & BCE & Average & Standard Deviation\\
\toprule[0.7pt]
\multicolumn{10}{c}{\textbf{Outex}}  \\ \hline  

FCNT & no & 65.38 & 62.84 & 72.91 & 72.66 & 77.96 & 60.09 & 68.64 & 11.70 \\
EWT-FCNT & no & 96.06 & 98.00 & 98.98 & 98.98 & 98.98 & 97.70 &98.12 & 1.22 \\ 
EWT-FCNT & yes & 96.52 & 98.27 & 99.13 & 99.13 & 99.13 & 98.02 & 98.36 & 0.74 \\

U-Net & yes & 72.83 & 64.25 & 71.40 & 70.85 & 80.47 & 61.89 & 70.28 & 14.86 \\
EWT-U-Net & yes & 97.51 & 98.79 & 99.39 & 99.39 & 99.39 & 98.54 & \textbf{98.83} & \textbf{0.73} \\
DA & yes & 70.70 & 60.65 & 67.77 & 67.27 & 78.75 & 58.58 & 67.28 & 16.06 \\
EWT-DA & yes & 95.84 & 97.82 & 98.89 & 98.89 & 98.89 & 97.45 & 97.97 & 0.89 \\
Siamese-Net  & yes & 67.46 & 56.77 & 64.36 & 63.60 & 76.46 & 53.80 & 63.74 & 12.89\\
EWT-Siamese-Net  & yes & 86.66 & 88.47 & 93.09 & 93.03 & 93.87 & 86.32 & 90.24 & 6.92\\
PSP-Net & yes & 86.00 & 84.22 & 88.00 & 87.83 & 91.64 & 82.63 & 86.72 & 14.10 \\
EWT-PSP-Net & yes & 96.05 & 97.93 & 98.94 & 98.94 & 98.94 & 97.53 & 98.06&1.04\\

\toprule[0.7pt]
\multicolumn{10}{c}{\textbf{UIUC}}  \\ \hline  
FCNT & no &  85.81 & 89.71 & 94.39 & 94.39 & 94.39 & 87.87 & 91.09 & 4.72 \\
EWT-FCNT & no & 93.32 & 95.88 & 97.85 & 97.85 & 97.85 & 95.05 & 96.30 & 2.65\\ 
EWT-FCNT & yes & 94.66 & 96.89 & 98.40 & 98.40 & 98.40 & 96.25 &  97.17 &  1.79\\ 
U-Net & yes & 88.73  & 90.62  & 94.72  & 94.63  & 94.97  & 88.64  & 92.05  & 5.73 \\
EWT-U-Net & yes & 95.87  & 97.50  & 98.71  & 98.71  & 98.71  & 96.84  & \textbf{97.72}  & 2.07 \\
DA & yes & 90.23  & 92.04  & 95.59  & 95.59  & 95.74  & 90.47  & 93.28  & 4.70 \\
EWT-DA & yes & 93.09  & 95.19  & 97.43  & 97.43  & 97.44  & 94.16  & 95.79  & 3.67 \\
Siamese-Net  & yes & 75.05  & 70.37  & 79.17  & 78.80  & 83.85  & 67.98  & 75.87  & 11.47 \\
EWT-Siamese-Net  & yes & 78.54  & 78.02  & 85.54  & 85.06  & 87.89  & 75.00  & 81.67  & 10.99 \\
PSP-Net & yes & 91.73  & 94.00  & 96.76  & 96.76  & 96.86  & 92.65  & 94.79  & 3.53 \\
EWT-PSP-Net & yes & 94.57  & 96.78  & 98.34  & 98.34  & 98.34  & 96.07  & 97.07&\textbf{1.64} \\
\bottomrule[0.7pt]
\end{tabular}
\end{threeparttable}
\end{table*}

\begin{figure}[!t]
\includegraphics[width=1\columnwidth]{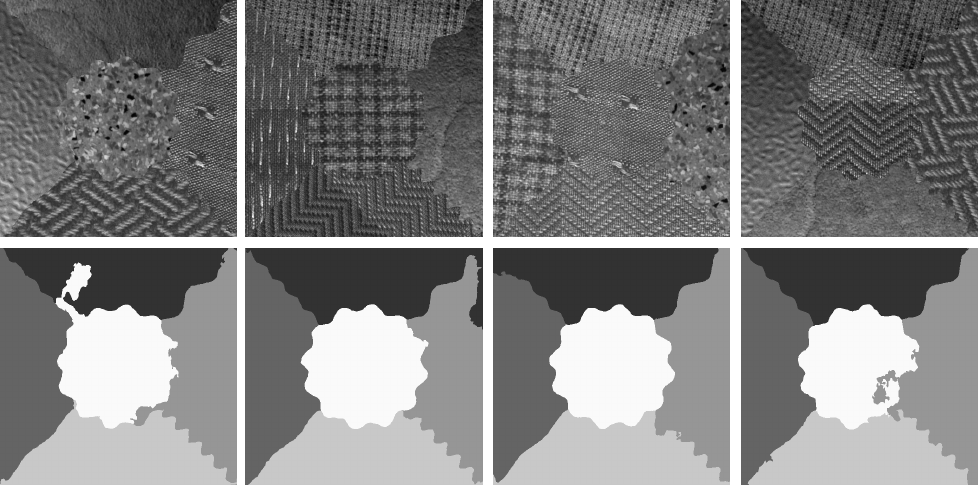}
\caption{Failure examples on the Outex dataset. First row: input test image to segment. Second row: EWT-U-Net segmentation results.}
\label{fig:outex_failure}
\end{figure}

Table~\ref{tab:g1} gives quantitative measurements for each dataset based on the previously mentioned metrics. We clearly observe that the proposed 
algorithm largely outperforms the FCNT and other CNN based approach. On average, our method EWT-FCNT reaches 98.36\% of good classification against 68.64\% for the classic FCNT on the Outex dataset and 97.17\% 
against 91.09\% on the UIUC dataset, respectively. The best results of both two datasets are EWT-U-Net which reaches 98.83\% of good classification versus 70.28\% for the classic U-Net on the Outex dataset and 97.72\% against 92.05\% on the UIUC dataset, respectively. The results in Table~\ref{tab:g1} also confirm that, whichever considered neural network, the addition of the empirical curvelet as texture descriptors permits to dramatically increase their efficacy. Moreover, the stability of our approach is also confirmed by the low value of the standard deviation. Figure \ref{fig:outex_failure} illustrate the four worst failure results obtained from the Outex dataset by the EWT-U-Net approach. From the benchmark results, we observe that these worst results still reach more than 96\% of good classification which again illustrates the efficacy of the proposed method. The corresponding computation costs for the Outex and UIUC test images for each method are available in Table \ref{tab:g2}. Except for the Siamese-Net architecture which uses a 32 $\times$ 32 sliding window, all other approaches spend very little time to segment the test images.

\begin{table*}[ht]
\centering
\scriptsize

\begin{threeparttable}
\caption{\label{tab:g2}Average computational speed in grayscale dataset for each image}

\begin{tabular}{ccccccccccc}
\toprule[0.7pt]
Method & FCNT & EWT-FCNT & UNet & EWT-UNet & DA & EWT-DA & SiameseNet & EWT-SiameseNet & PSPNet & EWT-PSPNet\\
\toprule[0.7pt]
Time & 3.03ms  & 152.82ms  & 0.068ms  & 154.55ms  & 2.93ms  & 153.18ms  & 101.79s  & 99.80s  & 0.076ms  & 166.22ms  \\

\bottomrule[0.7pt]
\end{tabular}
\end{threeparttable}

\end{table*}

\subsection{Color Textures}
\begin{figure}[!t]
\includegraphics[width=1\columnwidth]{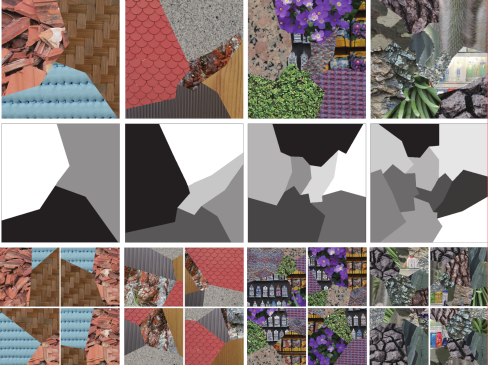}
\caption{Examples of test images from the Prague dataset (first row) and their corresponding ground-truth images (center row). The bottom row shows some generated training samples.}
\label{fig:prag_result}
\end{figure}

The Prague dataset \cite{Prague} consists of 20 texture mosaics which are synthetically generated from random compositions of 89 different pristine 
textures from 10 thematic categories. Each mosaic and its associated training textures are color images, they all have a size of $512 \times 512$ pixels. Some examples of
test input images and their ground-truths from that dataset are shown in the first and second rows of Fig.~\ref{fig:prag_result}, respectively. The number of classes on each 
image varies from 3 to 12 and a single training image is provided for each class. Quantitative comparisons are obtained, by using the measures provided on the Prague 
texture segmentation website \cite{Pragueweb}. These include: region-based metrics correct segmentation (CS), over-segmentation (OS), under-segmentation (US), missed 
error (ME), noise error (NE); pixel-based metrics: omission error (O), commission error (C), class accuracy (CA), recall (CO), precision (CC), type I error (I.), type II error 
(II.), mean class accuracy estimate (EA), mapping score (MS), root mean square proportion estimation error (RM), comparison index (CI); consistency-based metrics: global 
consistency
error (GCE) and local consistency error (LCE); and clustering: Mirkin metric (dM), Van Dongen metric (dD), the variation of information (dVI). We refer the reader to 
\cite{Pragueweb} for complete details and interpretations of these metrics. All these criteria, except dVI, are displayed after multiplication by 100.

Like for the grayscale experiments, we create a training set of a 1000 texture mosaics. To guarantee enough variability, we 
create mosaics of textures by randomly generating Voronoi polygons which are then filled with randomly selected pristine textures.  All training mosaics size is $512 \times 512$ 
pixels; some examples are illustrated in the bottom row of Fig.~\ref{fig:prag_result}. We experimentally found that, in the texture features extraction stage, a window size 
set to $19\times19$ pixels provides the best results. In addition, since the pristine color textures clearly contain both low and high frequencies, we kept the low frequency component extracted by the EWT.

\begin{figure*}[!t]
\begin{center}
\includegraphics[width=1.5\columnwidth]{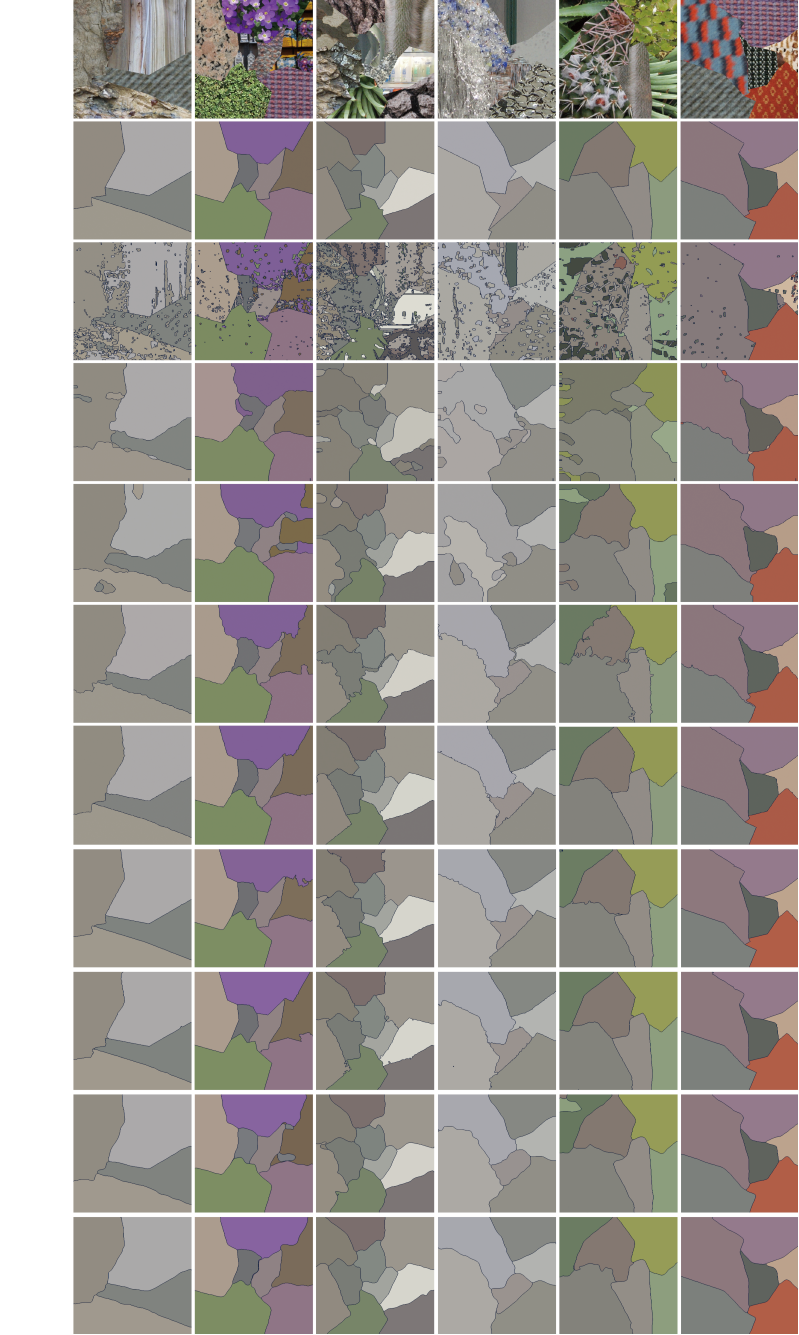}
\caption{Examples of segmentation results on Prague color texture segmentation dataset. From top to bottom: input test image, ground-truth, MRF, COF, Con-Col, FCNT, EWT-FCNT, U-Net, EWT-U-Net, PSP-Net, EWT-PSP-Net.}
\label{fig:prague_result}
\end{center}
\end{figure*} 

\begin{table*}[ht]
\centering
\scriptsize

\begin{threeparttable}

\caption{\label{tab:c1}Prague color dataset benchmark results. Each row corresponds to a segmentation quality measure and the arrow indicates whether large or low values are better. The first 
rank is marked by boldface and the second rank is marked by an asterisk. Here NR means no post-processing segmentation refinement has been performed. }

\begin{tabular}{lrrrrrrrrrrrrr}
\toprule[0.7pt]  

Method & MRF & COF & Con-Col & FCNT-NR & FCNT & EWT-FCNT-NR & EWT-FCNT & U-Net & EWT-U-Net & DA & EWT-DA & PSP-Net & EWT-PSP-Net\\ \midrule[0.5pt]  

$\uparrow$ CS  & 46.11 & 52.48 & 84.57 & 87.52 & 96.01 & 98.11* & \textbf{98.45} & 96.71 & 97.98 & 94.18 & 95.22 & 96.45 & 98.09\\ 
$\downarrow$ OS & 0.81 & \textbf{0.00} & \textbf{0.00} & \textbf{0.00} & 1.58 & \textbf{0.00} & \textbf{0.00} & 1.71 & 1.78 & \textbf{0.00} & 0.72 & 0.17* & 1.70\\ 
$\downarrow$ US & 4.18 & 1.94 & 1.70 & \textbf{0.00} & 1.20 & \textbf{0.00} & \textbf{0.00} & \textbf{0.00} & 0.23* & 1.18 & 2.61 & 0.41 & \textbf{0.00}\\
 $\downarrow$ ME & 44.82 & 41.55 & 9.50 & 6.70 & 0.78 & 0.49* & \textbf{0.37} & 0.68 & 0.78 & 3.42 & 2.76 & 1.23 & \textbf{0.37} \\ 
$\downarrow$ NE & 45.29 & 40.97 & 10.22 & 6.90 & 0.89 & 0.46* & 0.46* & 0.48 & 0.68 & 3.24 & 2.64 & 1.12 & \textbf{0.09}\\ 
$\downarrow$ O     & 14.52 & 20.74 & 7.00 & 7.49 & 2.72 & 1.16 & 0.93 & \textbf{0.72} & 0.78* & 3.13 & 0.92 & 2.75 & 1.57\\
 $\downarrow$ C   & 16.77 & 22.10 & 5.34 & 6.16 & 2.29 & 1.56 & 1.04* & \textbf{0.7} & 1.53 & 1.32 & 1.70 & 2.39 & 1.82\\
 $\uparrow$ CA & 65.42 & 67.01 & 86.21 & 87.08 & 93.95 & 97.01 & \textbf{97.67} & 95.86 & 97.24* & 94.53 & 95.45 & 93.89 & 96.52\\
 $\uparrow$ CO & 76.19 & 77.86 & 92.02 & 92.61 & 96.73 & 98.43* & \textbf{98.78} & 96.91 & 98.32 & 96.23 & 97.18 & 96.06 & 98.18\\
 $\uparrow$ CC & 80.30 & 78.34 & 92.68 & 93.26 & 97.02 & 98.46* & \textbf{98.81} & 97.38 & 98.4 & 97.01 & 97.78 & 96.41 & 98.01\\ 
$\downarrow$ I. & 23.81 & 22.14 & 7.98 & 7.39 & 3.27 & 1.57* & \textbf{1.22} & 3.09 & 1.68 & 3.77 & 2.82 & 3.94 & 1.82\\
 $\downarrow$ II. & 4.82 & 4.40 & 1.70 & 1.49 & 0.68 & 0.33 & 0.25* & 0.41 & \textbf{0.20} & 0.58 & 0.42 & 0.69 & 0.39\\
$\uparrow$ EA & 75.40 & 76.21 & 91.72 &   & 96.68 & 98.40* & \textbf{98.77} & 97.01 & 98.32 & 96.24 & 97.05 & 96.08 & 98.08\\
 $\uparrow$ MS & 64.29 & 66.79 & 88.03 &   & 95.10 & 97.65* & \textbf{98.17} & 95.37 & 97.49 & 94.35 & 95.76 & 94.08 & 97.27\\
 $\downarrow$ RM & 6.43 & 4.47 & 2.08 & 1.38 & 0.86 & 0.28* & \textbf{0.24} & 0.61 & 0.30 & 1.07 & 0.81 & 0.70 & 0.29\\
$\uparrow$ CI & 76.69 & 77.05 & 92.02 & 92.81 & 96.77 & 98.42* & \textbf{98.78} & 97.08 & 98.34 & 96.41 & 97.24 & 96.15 & 98.09\\
 $\downarrow$ GCE & 25.79 & 23.94 & 11.76 & 11.76 & 5.55 & 2.84 & 2.33 & \textbf{2.13} & 2.29* & 3.50 & 3.43 & 4.67 & 3.20\\
 $\downarrow$ LCE & 20.68 & 19.69 & 8.61 & 8.61 & 3.75 & 2.23 & 1.68 & \textbf{1.46} & 1.61* & 2.47 & 2.34 & 3.52 & 2.70\\
$\downarrow$ dD & 20.35 & 17.86 & 7.50 &   & 3.06 & 1.57 & \textbf{1.21} & 1.45 & 1.32* & 2.41 & 2.22 & 2.59 & 1.75 \\ 

$\downarrow$ dM & 13.25 & 10.62 & 4.69 &   & 1.96 & 0.99 & \textbf{0.74} & 0.77* & \textbf{0.74} & 1.35 & 1.26 & 1.56 & 1.08\\

$\downarrow$ dVI & 14.51 & 14.22 & 13.99 &   & 13.80 & 13.71 & \textbf{13.68} & \textbf{13.68} & 13.70* & 13.71 & 13.73 & 13.77 & 13.71\\
\bottomrule[0.7pt] 
\end{tabular}
\end{threeparttable}

\end{table*}

\begin{table*}[ht]
\centering
\footnotesize

\begin{threeparttable}
\caption{\label{tab:c2}Average computational speed on color datasets.}

\begin{tabular}{ccccccccc}
\toprule[0.7pt]
Method & FCNT & EWT-FCNT & U-Net & EWT-U-Net & DA & EWT-DA   & PSP-Net & EWT-PSP-Net\\
\toprule[0.7pt]
Time & 3.61ms  & 1.830s  & 4.98ms  & 1.833s  & 3.80ms  & 1.831s  & 14.39ms & 1.833s   \\

\bottomrule[0.7pt]
\end{tabular}
\end{threeparttable}

\end{table*}

Like in the previous section, we again compare the EWT-FCNT results (with and without refinement) with several neural networks (with and without feeding empirical curvelet features) and the three best algorithms reported on the Prague texture segmentation website \cite{Pragueweb}. These three algorithms are: 
1) the MRF algorithm based on a Markov Random Field pixel classification model \cite{MRF}, 2) the COF algorithm using co-occurrence 
features and the nearest neighbor classifier, 3) the Con-Col algorithm (no details are provided). Finally, the state-of-the-art results are achieved by the FCNT algorithm. Table~\ref {tab:c1} 
presents the benchmark results for these experiments. According to these metrics, our approach EWT-FCNT (with refinement) outperforms the state-of-the-art results indexed on the Prague texture segmentation website. Moreover, The EWT-U-Net, EWT-DA and EWT-PSP-Net approaches also provide better segmentation results than their original (i.e. with empirical curvelet) counterparts. It is again clear that the use of empirical curvelets to extract texture features to feed a 
neural network based segmentation algorithm dramatically impacts the segmentation performances. It is also noticeable that, for most experiments, even without the refinement step, the proposed algorithm EWT-FCNT outperforms most existing state-of-the-art approaches. Fig.~\ref{fig:prague_result} shows several segmentation results from the algorithms mentioned 
above for visual comparison. We can easily observe that our method not only gives the best visual segmentation (notably near the edges) but also provides the closest results to the ground-truths. The average computation costs of each datatest are presented in Table~\ref{tab:c2}.

\section{Conclusion}\label{sec:conc}
In this paper, we addressed the question of supervised texture segmentation/classification. We proposed a modified version of a Fully Convolutional Network where, as opposed to what was done in the current literature, we feed this network with specific adapted texture features. These texture features are obtained by building an ``optimized'' empirical curvelet filter bank. We experimentally verified that our proposed algorithm, tested on several classic datasets, outperforms all current state-of-the-art algorithms. Moreover, we also illustrate that the same strategy can be used for other neural network architectures and that it always dramatically improves the performances. These results tend  to show that convolutional layers in deep learning networks are not necessarily good at capturing the appropriate characteristics to perform accurate classification. This leads us to think that if some specific features exist for the type of data we have to classify, these features should be used to feed the used deep learning networks.

\section*{Acknowledgment}
This work was initiated during Yuan Huang stay at San Diego State University which was supported by the Chinese Scholarship Council. The authors want to thanks the anonymous reviewers for their comments which allowed to improve this manuscript.

\section*{References}



\bibliographystyle{elsarticle-num} 
\bibliography{yuan}

\begin{thebibliography}{10}
\expandafter\ifx\csname url\endcsname\relax
  \def\url#1{\texttt{#1}}\fi
\expandafter\ifx\csname urlprefix\endcsname\relax\def\urlprefix{URL }\fi
\expandafter\ifx\csname href\endcsname\relax
  \def\href#1#2{#2} \def\path#1{#1}\fi

\bibitem{tumor}
S.~A. Karkanis, D.~K. Iakovidis, D.~E. Maroulis, D.~A. Karras, M.~Tzivras, Computer-aided tumor detection in endoscopic video using color wavelet features, IEEE Transactions on Information Technology in Biomedicine 7~(3) (2003) 141--152.
\newblock \href {https://doi.org/10.1109/TITB.2003.813794} {\path{doi:10.1109/TITB.2003.813794}}.

\bibitem{ctd1}
T.~E. Boult, R.~J. Micheals, X.~Gao, M.~Eckmann, Into the woods: visual surveillance of noncooperative and camouflaged targets in complex outdoor settings, Proceedings of the IEEE 89~(10) (2001) 1382--1402.
\newblock \href {https://doi.org/10.1109/5.959337} {\path{doi:10.1109/5.959337}}.

\bibitem{ctd2}
P.~Bian, Y.~Jin, N.~Zhang, Fuzzy c-means clustering based digital camouflage pattern design and its evaluation, in: IEEE 10th International Conference on Signal Processing Proceedings, 2010, pp. 1017--1020.
\newblock \href {https://doi.org/10.1109/ICOSP.2010.5655856} {\path{doi:10.1109/ICOSP.2010.5655856}}.

\bibitem{micro}
Y.~Gao, M.~E. Helgeson, Texture analysis microscopy: quantifying structure in low-fidelity images of dense fluids, Opt. Express 22~(8) (2014) 10046--10063.
\newblock \href {https://doi.org/10.1364/OE.22.010046} {\path{doi:10.1364/OE.22.010046}}.

\bibitem{micro2}
C.~Jones, T.~Liu, M.~Ellisman, T.~Tasdizen, Semi-automatic neuron segmentation in electron microscopy images via sparse labeling, in: 2013 IEEE 10th International Symposium on Biomedical Imaging, 2013, pp. 1304--1307.
\newblock \href {https://doi.org/10.1109/ISBI.2013.6556771} {\path{doi:10.1109/ISBI.2013.6556771}}.

\bibitem{randomwalk}
J.~Shen, Y.~Du, W.~Wang, X.~Li, Lazy random walks for superpixel segmentation, IEEE Transactions on Image Processing 23~(4) (2014) 1451--1462.
\newblock \href {https://doi.org/10.1109/TIP.2014.2302892} {\path{doi:10.1109/TIP.2014.2302892}}.

\bibitem{superpix}
J.~Shen, X.~Hao, Z.~Liang, Y.~Liu, W.~Wang, L.~Shao, Real-time superpixel segmentation by {DBSCAN} clustering algorithm, IEEE Transactions on Image Processing 25~(12) (2016) 5933--5942.
\newblock \href {https://doi.org/10.1109/TIP.2016.2616302} {\path{doi:10.1109/TIP.2016.2616302}}.

\bibitem{Liu2019}
L.~Liu, J.~Chen, P.~Fieguth, G.~Zhao, R.~Chellappa, M.~Pietik{\"a}inen, From {BoW to CNN}: Two decades of texture representation for texture classification, International Journal of Computer Vision 127~(1) (2019) 74--109.
\newblock \href {https://doi.org/10.1007/s11263-018-1125-z} {\path{doi:10.1007/s11263-018-1125-z}}.

\bibitem{Tamura}
H.~Tamura, S.~Mori, T.~Yamawaki, Textural features corresponding to visual perception, IEEE Transactions on Systems, Man, and Cybernetics 8~(6) (1978) 460 -- 473.

\bibitem{Li2019}
N.~Li, Z.~Xiong, Automated brain tumor segmentation from multi-modality {MRI} data based on tamura texture feature and {SVM} model, Journal of Physics: Conference Series 1168 (2019) 032068.
\newblock \href {https://doi.org/10.1088/1742-6596/1168/3/032068} {\path{doi:10.1088/1742-6596/1168/3/032068}}.

\bibitem{Haralick}
R.~Haralick, Statistical and structural approaches to texture, Proceedings of the IEEE 67~(5) (1979) 786--804.

\bibitem{co-occ}
R.~M. Haralick, K.~Shanmugam, I.~Dinstein, Textural features for image classification, IEEE Transactions on Systems, Man, and Cybernetics SMC-3~(6) (1973) 610 -- 621.

\bibitem{Subudhi2018}
P.~Subudhi, authorSusanta Mukhopadhyay, A novel texture segmentation method based on co-occurrence energy-driven parametric active contour model, Signal, Image and Video Processing 12~(4) (2018) 669--676.
\newblock \href {https://doi.org/10.1007/s11760-017-1206-4} {\path{doi:10.1007/s11760-017-1206-4}}.

\bibitem{Zhao2018}
G.~Zhao, S.~Qin, D.~Wang, Interactive segmentation of texture image based on active contour model with local inverse difference moment feature, Multimedia Tools and Applications 77~(18) (2018) 24537–--24564.
\newblock \href {https://doi.org/10.1007/s11042-018-5777-z} {\path{doi:10.1007/s11042-018-5777-z}}.

\bibitem{Soares2018}
L.~D.~A. Soares, K.~F. C\^oco, E.~O.~T. Salles, P.~M. Ciarelli, Texture representation and classification with artificial hikers and fractals, in: IEEE Congress on Evolutionary Computation (CEC), Rio de Janeiro, Brazil, 2018.
\newblock \href {https://doi.org/10.1109/CEC.2018.8477788} {\path{doi:10.1109/CEC.2018.8477788}}.

\bibitem{Kunina2017}
I.~A. Kunina, L.~M. Teplyakov, A.~P. Gladkov, T.~M. Khanipov, D.~P. Nikolaev, Aerial images visual localization on a vector map using color-texture segmentation, in: Proc. SPIE 10696, Tenth International Conference on Machine Vision (ICMV 2017), Vienna, Austria, 2017, p. 106961T.
\newblock \href {https://doi.org/10.1117/12.2310138} {\path{doi:10.1117/12.2310138}}.

\bibitem{LBP}
T.~Ojala, M.~Pietikainen, T.~Maenpaa, Multiresolution gray-scale and rotation invariant texture classification with local binary patterns, IEEE Transactions on Pattern Analysis and Machine Intelligence 24~(7) (2002) 971 -- 987.

\bibitem{Wang2018}
T.~Wang, Y.~Dong, C.~Yang, L.~Wang, L.~Liang, L.~Zheng, J.~Pu, Jumping and refined local pattern for texture classification, IEEE Access 6 (2018) 64416--64426.
\newblock \href {https://doi.org/10.1109/ACCESS.2018.2877729} {\path{doi:10.1109/ACCESS.2018.2877729}}.

\bibitem{Banerjee2018}
P.~Banerjee, A.~K. Bhunia, A.~Bhattacharyya, P.~P. Roy, S.~Murala, Local neighborhood intensity pattern–a new texture feature descriptor for image retrieval, Expert Systems With Applications 113 (2018) 100--115.
\newblock \href {https://doi.org/10.1016/j.eswa.2018.06.044} {\path{doi:10.1016/j.eswa.2018.06.044}}.

\bibitem{Dong2018}
Y.~Dong, J.~Feng, C.~Yang, X.~Wang, L.~Zheng, J.~Pu, Multi-scale counting and difference representation for texture classification, The Visual Computer 34~(10) (2018) 1315–--1324.
\newblock \href {https://doi.org//10.1007/s00371-017-1415-4} {\path{doi:/10.1007/s00371-017-1415-4}}.

\bibitem{Kazak2018}
N.~Kazak, M.~Koc, Some variants of spiral lbp in texture recognition, IET Image Processing 12~(8) (2018) 1388--1393.
\newblock \href {https://doi.org/10.1049/iet-ipr.2017.1261} {\path{doi:10.1049/iet-ipr.2017.1261}}.

\bibitem{Chen2018}
W.~Chen, B.~Liao, W.~Li, Use of image texture analysis to find {DNA} sequence similarities, Journal of Theoretical Biology 455 (2018) 1--6.
\newblock \href {https://doi.org/10.1016/j.jtbi.2018.07.001} {\path{doi:10.1016/j.jtbi.2018.07.001}}.

\bibitem{Wardhani2018}
M.~K. Wardhani, X.~Yu, J.~Li, Statistical multi-scale laws’ texture energy for texture segmentation, in: Proc. SPIE 10828, Third International Workshop on Pattern Recognition, Jinan, China, 2018, p. 108280I.
\newblock \href {https://doi.org/10.1117/12.2501914} {\path{doi:10.1117/12.2501914}}.

\bibitem{Hermite}
L.~Debnath, On {Hermite} transform, Matemati\u cki Vesnik 1~(30) (1964) 285--292.

\bibitem{Ribas}
L.~C. Ribas, J.~J. M.~S. Junior, L.~F.~S. Scabini, O.~M. Bruno, \href{https://arxiv.org/pdf/1806.09170.pdf}{Fusion of complex networks and randomized neural networks for texture analysis}, preprint on ArXiV (2018).
\newline\urlprefix\url{https://arxiv.org/pdf/1806.09170.pdf}

\bibitem{UTSMRF}
B.~M. .~R. Chellappa, Unsupervised texture segmentation using markov random field models, IEEE Transactions on Pattern Analysis and Machine Intelligence 13~(5) (1991) 478 -- 482.

\bibitem{Lfilter}
T.~R. .~J. Husoy, Filtering for texture classification: a comparative study, IEEE Transactions on Pattern Analysis and Machine Intelligence 21~(4) (1999) 291 -- 310.

\bibitem{Ye2018}
Z.~Ye, X.~Hou, X.~Zhang, J.~Yang, Application of {Bat} algorithm for texture image classification, International Journal of Intelligent Systems and Applications 5 (2018) 42--50.
\newblock \href {https://doi.org/10.5815/ijisa.2018.05.05} {\path{doi:10.5815/ijisa.2018.05.05}}.

\bibitem{Song2018}
Y.~Song, S.~Zhang, B.~He, Q.~Sha, Y.~Shen, T.~Yan, R.~Nian, A.~Lendasse, Gaussian derivative models and ensemble extreme learning machine for texture image classification, Neurocomputing 277 (2018) 53--64.
\newblock \href {https://doi.org/10.1016/j.neucom.2017.01.113} {\path{doi:10.1016/j.neucom.2017.01.113}}.

\bibitem{Song2018a}
T.~Song, H.~Li, F.~Meng, Q.~Wu, J.~Cai, {LETRIST}: Locally encoded transform feature histogram for rotation-invariant texture classification, IEEE Transactions On Circuits And Systems For Video Technology 28~(7) (2018) 1565--1579.
\newblock \href {https://doi.org/10.1109/TCSVT.2017.2671899} {\path{doi:10.1109/TCSVT.2017.2671899}}.

\bibitem{Gabor1}
A.~K. Jain, F.~Farrokhnia, Unsupervised texture segmentation using {Gabor} filters, Pattern Recognition 24~(12) (1991) 1167--1186.

\bibitem{Gabor2}
D.~D. andW.E. Higgins, Optimal {Gabor} filters for texture segmentation, IEEE Transactions on Image Processing 4~(7) (1995) 947 -- 964.

\bibitem{Kim2018}
N.~C. Kim, H.~J. So, Directional statistical {Gabor} features for texture classification, Pattern Recognition Letters 112 (2018) 18--26.
\newblock \href {https://doi.org/10.1016/j.patrec.2018.05.010} {\path{doi:10.1016/j.patrec.2018.05.010}}.

\bibitem{Gao2019}
M.~Gao, H.~Chen, S.~Zheng, B.~Fang, Feature fusion and non-negative matrix factorization based active contours for texture segmentation, Signal Processing 159 (2019) 104--118.
\newblock \href {https://doi.org/10.1016/j.sigpro.2019.01.021} {\path{doi:10.1016/j.sigpro.2019.01.021}}.

\bibitem{wavelet1}
M.~Unser, Texture classification and segmentation using wavelet frames, IEEE Transactions on Image Processing 4~(11) (1995) 1549--1560.

\bibitem{wavelet2}
H.-C. Hsin, Texture segmentation using modulated wavelet transform, IEEE Transactions on Image Processing 9~(7) (2000) 1299 -- 1302.

\bibitem{wavelet3}
M.~Storath, A.~Weinmann, M.~Unser, Unsupervised texture segmentation using monogenic curvelets and the potts model, in: IEEE International Conference on Image Processing (ICIP), IEEE, Paris, France, 2014.

\bibitem{Subudhi2019}
P.~Subudhi, S.~Mukhopadhyay, An efficient graph reduction framework for interactive texture segmentation, Signal Processing: Image Communication 74 (2019) 42--53.
\newblock \href {https://doi.org/10.1016/j.image.2019.01.010} {\path{doi:10.1016/j.image.2019.01.010}}.

\bibitem{EMD}
N.~E. Huang, Z.~Shen, S.~R. Long, M.~C. Wu, H.~H. Shih, Q.~Zheng, N.-C. Yen, C.~C. Tung, H.~H. Liu, The empirical mode decomposition and the hilbert spectrum for nonlinear and non-stationary time series analysis, Proceedings of the Royal Society A Mathematical, Physical and Engineering Sciences.

\bibitem{Yang2018}
Z.~Yang, Q.~Zhang, F.~Zhou, L.~Yang, Hilbert spectrum analysis of piecewise stationary signals and its application to texture classification, Digital Signal Processing 82 (2018) 1--10.
\newblock \href {https://doi.org/10.1016/j.dsp.2018.07.020} {\path{doi:10.1016/j.dsp.2018.07.020}}.

\bibitem{EWT1D}
J.~Gilles, Empirical wavelet transform, IEEE Transactions on Signal Processing 61~(16) (2013) 3999--4010.

\bibitem{EWT2D}
J.~Gilles, G.~Tran, S.~Osher, {2D} empirical transforms. wavelets, ridgelets and curvelets revisited, SIAM Journal on Imaging Sciences 7~(1) (2014) 157--186.

\bibitem{UTSEWT}
Y.~Huang, V.~D. Bortoli, F.~Zhou, J.~Gilles, Review of wavelet-based unsupervised texture segmentation, advantage of adaptive wavelets, IET Image Processing.

\bibitem{AKBULUT2018494}
Y.~Akbulut, Y.~Guo, A.~Şeng\"ur, M.~Aslan, An effective color texture image segmentation algorithm based on {Hermite} transform, Applied Soft Computing 67 (2018) 494--504.
\newblock \href {https://doi.org/10.1016/j.asoc.2018.03.018} {\path{doi:10.1016/j.asoc.2018.03.018}}.

\bibitem{Meanshift}
D.~Comaniciu, P.~Meer, Mean shift: a robust approach toward feature space analysis, IEEE Transactions on Pattern Analysis and Machine Intelligence 24~(5) (2005) 603--619.

\bibitem{Pustelnik}
N.~Pustelnik, H.~Wendt, P.~Abry, N.~Dobigeon, Combining local regularity estimation and total variation optimization for scale-free texture segmentation, IEEE Transactions on Computational Imaging 2~(4) (2016) 468--479.
\newblock \href {https://doi.org/10.1109/TCI.2016.2594139} {\path{doi:10.1109/TCI.2016.2594139}}.

\bibitem{Pascal2018}
B.~Pascal, N.~Pustelnik, P.~Abry, M.~Serres, V.~Vidal, Joint estimation of local variance and local regularity for texture segmentation. application to multiphase flow characterization, in: 25th IEEE International Conference on Image Processing (ICIP), Athens, Greece, 2018.
\newblock \href {https://doi.org/10.1109/ICIP.2018.8451380} {\path{doi:10.1109/ICIP.2018.8451380}}.

\bibitem{neutrosophic}
D.~Koundal, Texture-based image segmentation using neutrosophic clustering, IET Image Processing 11~(8) (2017) 640--645.
\newblock \href {https://doi.org/10.1049/iet-ipr.2017.0046} {\path{doi:10.1049/iet-ipr.2017.0046}}.

\bibitem{LLIF}
M.~Kiechle, M.~Storath, A.~Weinmann, M.~Kleinsteuber, Model-based learning of local image features for unsupervised texture segmentation, IEEE Transactions on Image Processing 27~(4) (2018) 1994--2007.
\newblock \href {https://doi.org/10.1109/TIP.2018.2792904} {\path{doi:10.1109/TIP.2018.2792904}}.

\bibitem{enermin}
X.~Dong, J.~Shen, L.~Shao, M.-H. Yang, Interactive cosegmentation using global and local energy optimization, IEEE Transactions on Image Processing 24~(11) (2015) 3966--3977.
\newblock \href {https://doi.org/10.1109/TIP.2015.2456636} {\path{doi:10.1109/TIP.2015.2456636}}.

\bibitem{HOE1}
J.~Shen, J.~Peng, X.~Dong, L.~Shao, F.~Porikli, Higher order energies for image segmentation, IEEE Transactions on Image Processing 26~(10) (2017) 4911--4922.
\newblock \href {https://doi.org/10.1109/TIP.2017.2722691} {\path{doi:10.1109/TIP.2017.2722691}}.

\bibitem{HOE2}
J.~Peng, J.~Shen, X.~Li, High-order energies for stereo segmentation, IEEE Transactions on Cybernetics 46~(7) (2017) 1616--1627.
\newblock \href {https://doi.org/10.1109/TCYB.2015.2453091} {\path{doi:10.1109/TCYB.2015.2453091}}.

\bibitem{Ustyuzhaninov}
I.~Ustyuzhaninov, C.~Michaelis, W.~Brendel, M.~Bethge, \href{https://arxiv.org/abs/1807.02654}{One-shot texture segmentation}, preprint in ArXiV (July 2018).
\newline\urlprefix\url{https://arxiv.org/abs/1807.02654}

\bibitem{randomwalkmarkov}
X.~Dong, J.~Shen, L.~Shao, L.~V. Gool, Sub-{M}arkov random walk for image segmentation, IEEE Transactions on Image Processing 25~(2) (2016) 516--527.
\newblock \href {https://doi.org/10.1109/TIP.2015.2505184} {\path{doi:10.1109/TIP.2015.2505184}}.

\bibitem{Wang_2018_CVPR}
W.~Wang, J.~Shen, X.~Dong, A.~Borji, Salient object detection driven by fixation prediction, in: IEEE Conference on Computer Vision and Pattern Recognition, 2018.

\bibitem{videosaliency}
W.~Wang, J.~Shen, L.~Shao, Video salient object detection via fully convolutional networks, IEEE Transactions on Image Processing 27~(1) (2018) 38--49.
\newblock \href {https://doi.org/10.1109/TIP.2017.2754941} {\path{doi:10.1109/TIP.2017.2754941}}.

\bibitem{pyramid}
H.~Zhao, J.~Shi, X.~Qi, X.~Wang, J.~Jia, Pyramid scene parsing network, in: Conference on Computer Vision and Pattern Recognition, 2017.

\bibitem{Context}
H.~Zhang, K.~Dana, J.~Shi, Z.~Zhang, X.~Wang, A.~Tyagi, A.~Agrawal, Context encoding for semantic segmentation, in: IEEE Conference on Computer Vision and Pattern Recognition, 2018.

\bibitem{Hyperparameter}
X.~Dong, J.~Shen, W.~Wang, L.~Yu, L.~Shao, F.~Porikli, Hyperparameter optimization for tracking with continuous deep {Q-}learning, in: Conference on Computer Vision and Pattern Recognition, 2018.

\bibitem{Salhi2018}
K.~Salhi, E.~M. Jaara, M.~T. Alaoui, Texture image segmentation approach based on neural networks, International Journal of Recent Contributions from Engineering, Science \& IT (iJES) 6~(1).
\newblock \href {https://doi.org/10.3991/ijes.v6i1.8166} {\path{doi:10.3991/ijes.v6i1.8166}}.

\bibitem{MBLearning}
M.~Kiechle, M.~Storath, A.~Weinmann, M.~Kleinsteuber, Model-based learning of local image features for unsupervised texture segmentation, IEEE Transactions on Image Processing 27~(4) (2018) 1994 -- 2007.

\bibitem{CNN}
M.~Cimpoi, S.~Maji, A.~Vedaldi, Deep filter banks for texture recognition and segmentation, in: IEEE Conference on Computer Vision and Pattern Recognition (CVPR), IEEE, Boston, MA, USA, 2015.

\bibitem{CNN2}
V.~Andrearczyk, P.~F. Whelan, Using filter banks in convolutional neural networks for texture classification, Pattern Recognition Letters 84 (2016) 63 -- 69.
\newblock \href {https://doi.org/https://doi.org/10.1016/j.patrec.2016.08.016} {\path{doi:https://doi.org/10.1016/j.patrec.2016.08.016}}.

\bibitem{Andrearczyk2018}
V.~Andrearczyk, P.~F. Whelan, Convolutional neural network on three orthogonal planes for dynamic texture classification, Pattern Recognition 76 (2018) 36--49.
\newblock \href {https://doi.org/10.1016/j.patcog.2017.10.030} {\path{doi:10.1016/j.patcog.2017.10.030}}.

\bibitem{Xue2018}
J.~Xue, H.~Zhang, K.~Dana, Deep texture manifold for ground terrain recognition, in: Computer Vision and Pattern Recognition Conference, Salt Lake City, Utah, USA, 2018, pp. 558--567.

\bibitem{Fu2018}
C.~Fu, D.~Chen, E.~Delp, Z.~Liu, F.~Zhu, Texture segmentation based video compression using convolutional neural networks, in: IS\&T International Symposium on Electronic Imaging, Visual Information Processing and Communication IX, Society for Imaging Science and Technology, 2018, pp. 155--1--155--6.
\newblock \href {https://doi.org/10.2352/ISSN.2470-1173.2018.2.VIPC-155} {\path{doi:10.2352/ISSN.2470-1173.2018.2.VIPC-155}}.

\bibitem{LSTM}
W.~Byeon, T.~M. Breuel, Supervised texture segmentation using {2D LSTM} networks, in: IEEE International Conference on Image Processing (ICIP), IEEE, Paris, France, 2014.

\bibitem{Siamese2}
X.~Dong, J.~Shen, Triplet loss in {Siamese} network for object tracking, in: V.~Ferrari, M.~Hebert, C.~Sminchisescu, Y.~Weiss (Eds.), European Conference on Computer Vision, Springer International Publishing, 2018, pp. 472--488.
\newblock \href {https://doi.org/10.1007/978-3-030-01261-8\_28} {\path{doi:10.1007/978-3-030-01261-8\_28}}.

\bibitem{FullySiamese}
L.~Bertinetto, J.~Valmadre, J.~F. Henriques, A.~Vedaldi, P.~H.~S. Torr, \href{http://arxiv.org/abs/1606.09549}{Fully-convolutional {Siamese} networks for object tracking}, CoRR abs/1606.09549.
\newline\urlprefix\url{http://arxiv.org/abs/1606.09549}

\bibitem{Siamese}
R.~Yamada, H.~Ide, N.~Yudistira, T.~Kurita, Texture segmentation using {Siamese} network and hierarchical region merging, in: 24th International Conference on Pattern Recognition, 2018.

\bibitem{FCN}
E.~Shelhamer, J.~Long, T.~Darrell, Fully convolutional networks for semantic segmentation, IEEE Transactions on Pattern Analysis and Machine Intelligence 39~(4) (2017) 640--651.

\bibitem{FCNT}
V.~Andrearczyk, P.~F. Whelan, Texture segmentation with fully convolutional networks, arXiv preprint arXiv:1703.05230.

\bibitem{DeepVisual}
W.~Wang, J.~Shen, Deep visual attention prediction, IEEE Transactions on Image Processing 27~(5).

\bibitem{vnet}
F.~Milletari, N.~Navab, S.-A. Ahmadi, {V-Net}: Fully convolutional neural networks for volumetric medical image segmentation, in: Fourth International Conference on 3D Vision, 2016.

\bibitem{unet}
O.~Ronneberger, P.~Fischer, T.~Brox, U-net: Convolutional networks for biomedical image segmentation, in: N.~Navab, J.~Hornegger, W.~M. Wells, A.~F. Frangi (Eds.), Medical Image Computing and Computer-Assisted Intervention -- MICCAI 2015, Springer International Publishing, Cham, 2015, pp. 234--241.

\bibitem{scalespace}
J.~Gilles, K.~Heal, A parameterless scale-space approach to find meaningful modes in histograms - application to image and spectrum segmentation, International Journal of Wavelets, Multiresolution and Information Processing 12~(6) (2014) 1450044--1--1450044--17.
\newblock \href {https://doi.org/10.1142/S0219691314500441} {\path{doi:10.1142/S0219691314500441}}.

\bibitem{Whitening}
Z.~Li, Y.~Fan, W.~Liu, The effect of whitening transformation on pooling operations in convolutional autoencoders, EURASIP Journal on Advances in Signal Processing 2015~(1) (2015) 37.
\newblock \href {https://doi.org/10.1186/s13634-015-0222-1} {\path{doi:10.1186/s13634-015-0222-1}}.

\bibitem{ZCA}
A.~J. Bell, T.~J. Sejnowski, The “independent components” of natural scenes are edge filters, Vision Research 37~(23) (1997) 3327--3338.
\newblock \href {https://doi.org/10.1016/S0042-6989(97)00121-1} {\path{doi:10.1016/S0042-6989(97)00121-1}}.

\bibitem{Outex}
T.~Ojala, T.~M{\"a}enp{\"a}{\"a}, M.~Pietik{\"a}inen, J.~Viertola, J.~Kyll{\"o}nen, S.~Huovinen, Outex - new framework for empirical evaluation of texture analysis algorithms., 2002, proc. 16th International Conference on Pattern Recognition, Quebec, Canada, 1:701 - 706.

\bibitem{UIUC}
S.~Lazebnik, C.~Schmid, J.~Ponce., A sparse texture representation using local affine regions., IEEE Transactions on Pattern Analysis and Machine Intelligence 27~(8) (2005) 1265--1278.

\bibitem{Prague}
M.~Haindl, S.~Mikes, Texture segmentation benchmark, in: 19th International Conference on Pattern Recognition (ICPR), IEEE, 2008.

\bibitem{benchmark}
J.~Pont-Tuset, F.~Marques, Measures and meta-measures for the supervised evaluation of image segmentation, in: IEEE Conference on Computer Vision and Pattern Recognition (CVPR), IEEE, 2013.

\bibitem{Pragueweb}
{Prague texture datasets}, \url{http://mosaic.utia.cas.cz}, accessed: 2018-06-11.

\bibitem{MRF}
Z.~Kato, T.-C. Pong, J.~C.-M. Lee, Color image segmentation and parameter estimation in a markovian framework, Pattern Recognition Letters 22~(3) (2001) 309 -- 321.
\newblock \href {https://doi.org/https://doi.org/10.1016/S0167-8655(00)00106-9} {\path{doi:https://doi.org/10.1016/S0167-8655(00)00106-9}}.

\end{thebibliography}





\end{document}